%% file: main.tex
\documentclass[journal]{IEEEtran}

\usepackage{cite}

\usepackage{hyperref}

\usepackage{booktabs}
\usepackage{multirow}
\usepackage{tablefootnote}
\usepackage{array}
\usepackage{authblk}
\usepackage{xspace}
\usepackage[table]{xcolor}
\usepackage{float}

\ifCLASSINFOpdf
  \usepackage[pdftex]{graphicx}
\else
\fi
\usepackage{amsfonts, amsmath, amssymb}
\usepackage{pifont}
\usepackage{amsmath}
\usepackage{bbold}
\usepackage[nameinlink,capitalize]{cleveref}

\usepackage[]{mdframed}
\usepackage{algpseudocode}

\usepackage{url}

\usepackage{xr-hyper}
\input{commands}

\begin{document}
%
\title{Regression-aware Continual Learning for Android Malware Detection}
%
%
%
\author[1,2]{Daniele Ghiani}
\author[1,4]{Daniele Angioni}
\author[1]{Giorgio Piras}
\author[1]{Angelo Sotgiu}
\author[1]{Luca Minnei}
\author[1,2]{Srishti Gupta}
\author[1]{Maura Pintor,~\IEEEmembership{Member,~IEEE}}
\author[1,3,4]{Fabio Roli~\IEEEmembership{Fellow,~IEEE}}
\author[1,4]{Battista Biggio,~\IEEEmembership{Fellow,~IEEE}}

\affil[1]{Department of Electrical and Electronic Engineering,
University of Cagliari, Italy}
\affil[2]{Department of Computer, Control and Management Engineering, Sapienza University, Rome, Italy}
\affil[3]{Department of Informatics, Bioengineering, Robotics, and Systems Engineering, University of Genoa, Italy}
\affil[4]{CINI, Italy}

\renewcommand{\subsectionautorefname}{Sect.}
\renewcommand{\sectionautorefname}{Sect.}
\renewcommand{\figureautorefname}{Fig.}


\maketitle

\begin{abstract}
Malware evolves rapidly, forcing machine learning-based detectors to be continuously updated. With antivirus vendors processing hundreds of thousands of new samples daily, datasets can grow to billions of examples, making full retraining impractical. 
\textit{Continual learning} (CL) has emerged as a scalable alternative, enabling incremental updates without full data access while mitigating \textit{catastrophic forgetting}.
In this work, we analyze a critical yet overlooked issue in this context: \textit{security regression}. Unlike forgetting, which manifests as a drop in average performance on previously seen data, security regression captures harmful sample-level prediction changes, e.g., malware samples that were correctly detected before an update but evade detection afterward.
This poses serious risks in security-critical applications, as the silent reintroduction of previously detected threats may undermine users’ trust in the update process, leading them to perceive a regression in security even if the average model performance has actually improved.
We first formalize and quantify security regression in CL-based malware detectors, revealing that up to 3-6\% of malware experience it after model updates.
We then address this issue by introducing a regression-aware framework to the CL setting. Specifically, we instantiate it via Positive Congruent Training (PCT), showing seamless integration with any prior CL strategy.
Experiments on the ELSA, Tesseract, and AZ-Class datasets show that our method effectively halves regression across different CL scenarios while maintaining strong detection performance over time.
\end{abstract}


%
\IEEEpeerreviewmaketitle

\section{Introduction} \label{intro}
%
The detection of malicious software (\aka \textit{malware}) is a central aspect, as it helps prevent the exploitation of software vulnerabilities and protects the integrity of the entire software security chain.
Malware detection is nowadays commonly entrusted to Machine Learning (ML) models, which, by leveraging large amounts of data, are able to learn complex correlations and patterns, ultimately enabling automatic detection~\cite{arp2014drebin, zhang2020enhancing
}.
However, ML models are primarily designed to function under the assumption of stationary data distributions, i.e., data that remain consistent over time.
This assumption does not hold in the ``wild'', where detectors can be impacted by a phenomenon known as \textit{concept drift}, \ie a change in the data distribution over time.
%
In particular, existing malware is progressively modified to evade detection, while the emergence of new malware families further challenges models' predictions and security measures by introducing completely new patterns.
Legitimate applications (\aka \textit{goodware}) are updated as well, but this typically has a minimal effect on the detector's performance~\cite{pendlebury2019tesseract, angioni2022itasec}.
Due to this drift, ML-based detectors suffer from rapid performance degradation over time~\cite{pendlebury2019tesseract}.
To address this degradation, the model must be regularly updated with new data.
One option is to retrain it using both newly collected and previous samples.
However, this quickly becomes infeasible as the dataset grows over time. 
For example, the AV-TEST Institute observes $\sim 450,000$ new malware samples each day, while VirusTotal processes over a million files daily.\footnote{\url{https://www.av-test.org/en/}, \url{https://www.virustotal.com/gui/}}
An alternative approach is to retrain the detector with only newly collected data. 
However, this often causes the resulting detector to entirely prioritize adaptation to recent distributions, at the cost of disregarding ``previously acquired knowledge''.
This issue is commonly known as \textit{catastrophic forgetting}~\cite{de2021continual}.
In this regard, Continual Learning (CL) addresses the forgetting problem, \ie preserving performance on previous data~\cite{de2021continual}.
While multiple CL methods exist~\cite{de2021continual}, only a few works apply them to malware detection~\cite{rahman2022limitations, rahman2025madarefficientcontinuallearning, kou2023sscl, park2025malcl}.

In addition to catastrophic forgetting, model updates in CL scenarios can lead to a distinct and overlooked issue known as \textit{regression}~\cite{yan2021positive}.
This phenomenon refers to a degradation in the model’s behavior on specific individual samples, particularly those that were correctly classified before an update but are misclassified afterward.
Unlike forgetting, regression can occur even when the overall performance of the new model improves: an updated model may report better detection while introducing regression on previously well-learned distributions. 
This is because train and update procedures are typically designed to minimize an aggregate loss, with no explicit constraints to preserve the model's predictions on specific samples that were already correct.
In security applications, this issue is particularly severe, as model updates may cause previously detected malware samples to evade detection again~\cite{Botacin2020we}, thereby undermining user trust and creating the perception of a security drop.\footnote{\url{https://cloud.google.com/blog/topics/threat-intelligence/churning-out-machine-learning-models-handling-changes-in-model-predictions/}}
We refer to this phenomenon as \textit{security regression}.
While recent work addresses regression on computer vision and natural language domains~\cite{yan2021positive, zhao2024elodi}, they: (i) overlook security-related tasks such as malware detection; and (ii) do not consider CL settings, where the lack of past data further complicates retaining prediction consistency.

In this work, we fill this gap by analyzing the security regression phenomenon of CL strategies for malware detection. 
Building on our initial findings in~\cite{ghiani2025understanding},\footnote{While~\cite{ghiani2025understanding} consists of an initial exploration of the regression phenomena in CL for malware detection, in this work we provide a consolidated and significantly extended treatment, supported by a broader experimental evaluation and deeper analysis.} 
we formalize this phenomenon in the context of CL for malware detection, and introduce a regression-aware training framework that can be integrated into existing CL strategies.
As an instantiation of this framework, we employ \textit{Positive Congruent Training} (PCT)~\cite{yan2021positive}, which jointly minimizes the classification loss and a regularization term designed to mitigate regression.
%
We evaluate our approach on three Android malware datasets (\elsa, \tesseract, and \azclass), considering five representative CL strategies and four different state-of-the-art malware detectors, each obtained as a combination of Drebin~\cite{arp2014drebin} or APIGraph~\cite{zhang2020enhancing} feature extractors with linear SVM or MLP models.
Our analysis shows that incrementally updating ML models with CL strategies introduces security regression, i.e., up to 3–6\% of previously detected malware samples are misclassified after model updates.
Integrating PCT into these strategies consistently mitigates this issue, reducing regression by up to 50\% across datasets and scenarios while maintaining competitive detection performance in most cases.
%
%

We summarize our contributions as follows:
(i) we formalize the notion of security regression in CL for malware detection as an additional side effect besides catastrophic forgetting;
(ii) we show empirically that security regression can occur even when forgetting is mitigated, affecting up to 3-6\% of previously detected malware after model updates; and
(iii) we introduce a regression-aware training framework instantiated via Positive Congruent Training (PCT) that can be integrated into existing CL pipelines, mitigating regression across datasets, detectors, and CL strategies.

\section{Continual Learning for Malware Detection}\label{sect:soa_cl}
CL algorithms update ML models with new data while addressing \textit{catastrophic forgetting}, \ie forgetting knowledge from past data.
In this section, we discuss CL scenarios and strategies, with a specific focus on malware detection.

\myparagraph{Learning Incrementally.}
CL can be formalized as learning from a sequence of $K$ sets of data samples $\{\mathcal{D}_k\}_{k=1}^K$, called \textit{experiences}, assuming no access to past data when learning new experiences.
%
Each experience consists of a dataset $\mathcal{D}_k=\{\vct x_i^k, y_i^k\}_{i=1}^{n^k}$, where $\vct x_{i}^{k} \in \set{X}^k$ is an input sample, $y_{i}^{k} \in \set{Y}^k$ its ground-truth label, and $n^k$ is the number of samples in the $k$-th experience.
The CL literature distinguishes three main settings: (i) \textit{domain-incremental}, (ii) \textit{class-incremental}, and \textit{task-incremental} learning~\cite{van2022three}.
In \textit{domain-incremental learning} (DIL), the input distribution varies across experiences, \ie $p(\vct{x}^k) \neq p(\vct{x}^{k+1})$, while the label space remains fixed, that is $\mathcal{Y}^k = \mathcal{Y}^{k+1}$.
In \textit{class-incremental learning} (CIL), the change affects the label space, hence $\mathcal{Y}^k \ne \mathcal{Y}^{k+1}$. 
Each experience $\mathcal{D}_k$ introduces new class labels, expanding the set of observed classes, $\hat{\set{Y}}^{k} = \cup_{j=1}^k \set{Y}^j$.
While the shift originates in the label space, the posterior distribution $p(\vct x | y)$ for the new classes is also influenced, making it more challenging than DIL.
In \textit{task-incremental learning} (TIL), each experience requires task labels at test time, which is unrealistic for malware detection. For this reason, it is not considered in this study.

\myparagraph{Mitigating Catastrophic Forgetting.}
As the model optimizes its parameters based on the current experience $\mathcal{D}_k$, it overrides knowledge from previous experiences $\mathcal{D}_j$, with $j<k$, causing forgetting. Forgetting after the $k$-th update is measured as: 
\begin{equation}
    F^k_j = \max_{o \in \{1,\dots,k-1\}} \mathcal{A}_j^o - \mathcal{A}_j^k, \quad \forall j<k \, ,
    \label{eq:forgetting_base}
\end{equation}
where $\mathcal{A}$ is a performance metric (typically the accuracy). 
Given a model trained on the $k$-th training experience, the first term on the right-hand side of the equation indicates the best performance among old updates on the experience $j$, while the second term indicates the performance of the model trained on the $k$-th experience but evaluated on the experience $j$.

CL strategies address forgetting by learning, at each experience $k$, an optimal parametrization $\vct{\theta}_k$ that generalizes to both current and past data, despite limited or no access to previous training experiences.
They are typically grounded into three approaches~\cite{de2021continual}: (i) \textit{replay-based}, (ii) \textit{regularization-based}, or (iii) \textit{parameter isolation-based}.
\textit{Replay methods} 
alleviate forgetting by either storing a subset of past data or generating pseudo-samples with a generative model. 
These samples are then replayed during training on new experiences, either to reinforce previous knowledge through rehearsal~\cite{rolnick2019experience} or to guide the optimization process in a way that minimizes interference with past tasks, as in Average-Gradient Episodic Memory (A-GEM~\cite{chaudhry2018efficient}).
\textit{Regularization methods} reduce memory requirements and prioritize privacy by avoiding storing past data.
They introduce additional loss terms constraining the update of parameters critical for past experiences.
For instance, Elastic Weight Consolidation (EWC~\cite{kirkpatrick2017overcoming}) and Synaptic Intelligence (SI~\cite{zenke2017continual}) estimate the importance of each parameter and penalize their deviation from previous optima.
\textit{Parameter-isolation methods} prevent forgetting by dedicating different model parameters to each experience~\cite{rusu2016progressive, xu2018reinforced}.
Since the latter methods are mostly used for TIL scenarios, we focus only on replay and regularization methods.

\myparagraph{Framing CL for Malware Detection.}
We frame the DIL and CIL scenarios in the context of malware classification.
In DIL, the input distribution evolves over time, while the label space remains fixed.
This is the case for the changes in the benign/malicious software landscape, while the classification task still remains to discern malware from goodware.
Given that applications are typically associated with a timestamp $t$ describing the first appearance date, we can describe concept drift, similarly to~\cite{pendlebury2019tesseract}, by defining the $k$-th experience $\mathcal{D}_k = \{\vct{x}_i^k, y_i^k, t_i^k\}_{i=1}^{n^k}$ over a time interval $\Delta t^k = [t^k, t^{k+1})$, where $t_i^k$ denotes the timestamp of each sample.
In CIL, the label space expands across experiences, with each experience introducing a distinct set of classes.
In the malware domain, this corresponds to the fine-grained task of distinguishing \textit{malware families}.
In this case, the model must learn new families while retaining performance on previous ones.
Formally, the label set grows over time such as $\mathcal{Y}^k \subset \mathcal{Y}^{k+1}$, and the label distribution shifts accordingly: $P(\mathcal{Y}^k) \ne P(\mathcal{Y}^{k+1})$.

\section{Security Regression in Continual Learning}\label{sect:soa_reg}
Although CL mitigates forgetting, detector updates can lead to fine-grained performance degradation in previously correctly classified samples, known as \textit{regression}~\cite{yan2021positive}, which we frame in the malware detection context as \textit{security regression}. 

\subsection{Security Regression}
\label{subsect:sec_reg}
Security regression is particularly problematic in malware detection, as reopening previously patched vulnerabilities can have serious consequences. 
%
In fact, while forgetting accounts for an overall degradation in detector performance, operating in such scenarios imposes special attention on previously detected malware samples that, after updates, bypass detection systems.
This is particularly relevant in the Android domain, where a relatively small number of families dominate the scene for several years, continuously generating new polymorphic variants~\cite{haque2025lamda,sun2025temporal}.
Within this temporal horizon, old malware samples remain representative of real-world threats, as they can be reused or slightly modified rather than replaced.
Antivirus systems often combine signature-based detection with ML: while signatures guarantee stable predictions for previously observed malware, they fail when faced with slight modifications~\cite{faruki2014android, canfora2015obfuscation}, making ML-based detectors essential for generalizing beyond exact signatures.
As these models are periodically updated, changes in predictions across model versions may cause security regression.
Thus, quantifying and addressing such inconsistencies becomes crucial.

\myparagraph{Formal Definition.}
To formally characterize security regression, we rely on the general notion of \textit{negative flips} (NFs) provided in~\cite{yan2021positive}, which captures prediction inconsistencies between model updates.
Let $f_{old}$ and $f_{new}$ denote, respectively, a model before and after an update, and let $(\vct x,y)$ be a sample with its ground truth label.
A NF occurs if $f_{old}(x) = y$ and $f_{new}(x) \neq y$, i.e., a shift from correct to incorrect classification of the same sample after the update.
Being $\Ind$ the indicator function, the fraction of NFs on a dataset $\set D$ of $N$ samples is referred to as \textit{negative flip rate} (NFR):
\begin{equation}
\text{NFR} = \frac{1}{N} \sum_{i=1}^N \Ind (f_{new}(x_i) \neq y_i \land f_{old}(x_i) = y_i).
\label{eq:nfr}
\end{equation}
%


\subsection{Security Regression in CL}
\label{subsect:sec_reg_cl}
In a CL scenario, the model is incrementally updated over a sequence of experiences $\{\mathcal{D}_k\}_{k=1}^K$.
Each update yields a new model parameterization, resulting in a sequence of models $\{f_k\}_{k=1}^K$, where $f_k$ denotes the model after training on experience $\mathcal{D}_k$.
To quantify \textit{security regression} across updates, we can adapt \cref{eq:nfr} to the CL case as follows:
\begin{equation}
\text{NFR}^k_j = \frac{1}{n^j} \sum_{i=1}^{n^j} \Ind (f_{k}(x_i) \neq y_i \land f_{k-1}(x_i) = y_i) \,
\label{eq:nfr_cl}
\end{equation}
where $\text{NFR}_j^k$ denotes the NFR for the $k$-th update $f_{k-1} \rightarrow f_k$ computed on the $j$-th experience composed by $n^j$ samples. 
This allows comparing the predictions of the newly updated model $f_k$ with its previous version $f_{k-1}$ on a selected test experience $\set{D}_j$, and measuring the fraction of samples that were correctly classified by $f_{k-1}$ but misclassified by $f_k$.
We evaluate $\text{NFR}_j^k$ after learning the $k$-th experience according to two modes: (i) the \textit{backward}, and (ii) the \textit{forward mode}.

\myparagraph{Backward Mode.} 
NFR in backward mode measures how the newly updated model $f_k$ performs on earlier experiences compared to its predecessor $f_{k-1}$, and is computed as:

\begin{equation}
\text{NFR}_B^k = \frac{1}{k'} \sum_{j=1}^{k'} \text{NFR}_j^k \ ,
\label{eq:nfr_backward}
\end{equation}
where $k'=k$ for DIL, and $k'=k-1$ for CIL, to ensure evaluation is restricted to classes known by both $f_k$ and $f_{k-1}$.
This aligns with standard protocols for measuring forgetting, tracking degradation on previously seen data after each update.
We use this setting to measure \textit{security regression} on past data. 

\myparagraph{Forward Mode.} NFR in forward mode compares the updated model $f_k$ to $f_{k-1}$, on data from the current and future experiences, highlighting early signs of degraded generalization to unseen future data. The forward NFR is computed as:
\begin{equation}
\text{NFR}_F^k = \frac{1}{(K-k+1)} \sum_{j=k}^K \text{NFR}_j^k \ .
\label{eq:nfr_forward}
\end{equation}
Forward evaluation is specific to DIL, where classes are fixed, and input distributions drift. 
We employ it to measure models' adaptability to future data distributions drifting over time.


\subsection{Forgetting vs. Regression} \label{subsect:reg_vs_forg}
\begin{figure*}[t]
    \centering
    \includegraphics[width=0.99\linewidth]{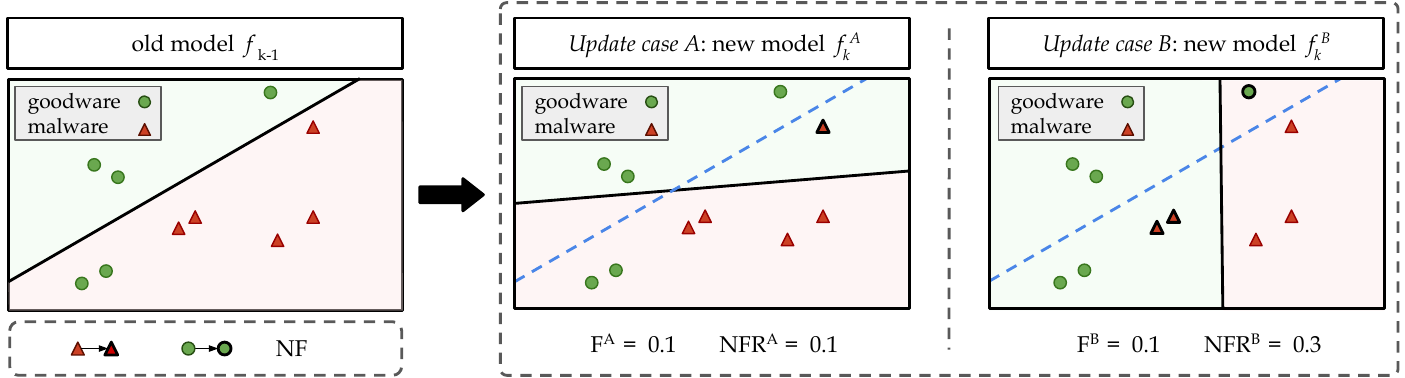}
    \caption{
    Forgetting vs. Regression. Starting from the old model $f_{k-1}$ (\textit{left}), we show two updates, $f_k^A$ and $f_k^B$ (\textit{right}), with identical forgetting ($\mathcal{F}=\text{NFR}-\text{PFR}$) but different regression: case A has 1/10 negative flips, case B has 3/10.
    }
    \label{fig:forg_vs_reg}
\end{figure*}
We now analyze its relation and difference with the forgetting phenomenon and highlight why both should be considered when assessing update quality.
Forgetting refers to a decline in model performance on previous experiences after the update.
Unlike forgetting, regression captures fine-grained degradations in model behavior, thus capturing instances where the model regresses in specific predictions (\ie focusing on negative flips), even when the overall performance improves.
Therefore, regression can be defined as a sample-wise measure, rather than an average, global one, as it provides a fine-grained assessment of prediction consistency over time.
A model might not exhibit forgetting and even improve the overall performance while still suffering from regression.

To formally analyze the relation between forgetting and regression, we first define the Positive Flip Rate (PFR), which follows a similar rationale to the NFR definition in \cref{eq:nfr_cl}:
\begin{equation}
    \text{PFR}^k_j = \frac{1}{n^j} \sum_{i=1}^{n^j} \Ind (f_{k}(x_i) = y_i \land f_{k-1}(x_i) \neq y_i) \, ,
    \label{eq:pfr_cl}
\end{equation}
where, conversely to NFR, PFR quantifies the percentage of samples previously misclassified by $f_{old}$ and that the new model $f_{new}$ corrected.
Let us now focus on the model update $f_{k-1} \rightarrow f_k$, evaluated on the experience $j=k-1$ (in the following, we omit the subscripts to improve readability).
Assuming that $f^{k-1}$ is the best-performing model among the previous ones (hence removing the $\max$ operator), the forgetting in \cref{eq:forgetting_base} takes the form of a simple difference in accuracy between the two considered models.
We can thus express the accuracy of model $k$ through the regression notation, \ie as the balance between NFR and PFR, yielding:
%
\begin{align}
F^k &= \mathcal{A}^{k-1} - \mathcal{A}^k \nonumber = \mathcal{A}^{k-1} - (\mathcal{A}^{k-1} - \text{NFR}^k + \text{PFR}^k) \nonumber \\
  &= \text{NFR}^k - \text{PFR}^k
\label{eq:forgetting_pfnf}
\end{align}
%
Hence, forgetting reflects the balance of gains and losses, while regression focuses solely on degradations and can occur even when forgetting is zero.
Clearly, this is particularly relevant in security-critical tasks, where negative shifts can have disproportionate consequences. 
We provide an overview of such a difference in \autoref{fig:forg_vs_reg}, illustrating a toy example with two update scenarios that share the same overall forgetting but differ significantly in terms of regression.

\section{Mitigating Security Regression}
\label{sect:reg_mitigation}
To mitigate security regression, we leverage Positive Congruent Training (PCT)~\cite{yan2021positive}.
We first describe PCT and then show how it can be integrated into CL strategies for Android malware detection.

\myparagraph{Positive Congruent Training (PCT).} PCT aims to find a model $f$ with reduced regression from a previous parameterization $f_{old}$. 
This is achieved by penalizing incorrect prediction changes of single samples through a distillation-like regularization term, $\mathcal{L}_{PC}$, which is added to the training loss $\mathcal{L}$ (\eg the cross-entropy loss):
\begin{equation}
\min_{f \in \set{F}} \, \sum_{i=1}^N \mathcal{L}(y_i, \vct{x}_i) + \lambda \mathcal{L}_{PC}(f(\vct{x}_i), f_{old}(\vct{x}_i)),
\end{equation}
where $\mathcal{L}_{PC}$ penalizes changes in the decision function regions containing samples that $f_{old}$ already classified correctly, while $\lambda$ is a hyperparameter controlling the regularization strength.
The term $\mathcal{L}_{PC}$ is the \textit{focal distillation}, defined as:
\begin{equation}
\mathcal{L}_{PC} = \sum_{i=1}^N \left[ \alpha + \beta \cdot \Ind\left( {f}_{old}(x_i) = y_i \right) \right] \mathcal{L}_D \left(f(\vct{x}_i), f_{old}(\vct{x}_i) \right),
\label{eq:PCT}
\end{equation}
where $\Ind$ is the indicator function, $\alpha$ is a base factor applied to all samples in the training set, while $\beta$ is a factor weighting samples correctly predicted by the old model, and $\mathcal{L}_D$ is a generic knowledge distillation loss.
In practice, we employ the loss denoted as \textit{focal distillation with logit matching} (FD-LM) in ~\cite{yan2021positive}, which can be formulated as the squared Euclidean distance between the outputs of $f$ and $f_{old}$ as follows:
\begin{equation}
\mathcal{L}_D(f(\vct{x}), f_{old}(\vct{x})) = \frac{1}{2} \left\| f(\vct{x}) - f_{old}(\vct{x}) \right\|_2^2.
\label{eq:dist}
\end{equation}

\myparagraph{Regression-aware CL through PCT.}
%
Mitigating regression in CL strategies amounts to adding the PCT regularization term $\mathcal{L}_{PC}$ described in \cref{eq:PCT}, scaled through a hyperparameter $\lambda$, to the training loss. 
%
This extension is agnostic of the CL strategy and promotes prediction consistency on previously correct samples without altering the CL formulation, making it a versatile plug-in against regression.
%
%
For replay methods, $\lambda \mathcal{L}_{PC}$ can directly be added to the training objective. 
In regularization methods, the CL strategy already consists of a regularization term. In this case, $\lambda \mathcal{L}_{PC}$ is added to the training objective and jointly optimized. 
We note that the proposed framework can be seamlessly instantiated with alternative non-regression penalties, such as the ones proposed in~\cite{zhao2024elodi, angioni2024robustnesscongruent}.

\section{Experiments}
\label{exp}
Our experiments aim to quantify the \textit{security regression} affecting CL algorithms in malware detection, and to demonstrate how this issue can be addressed via PCT regularization.

\subsection{Experimental Setup}
\label{subsect:exp_setup}

\myparagraph{DIL Scenario.}
In malware detection, a DIL scenario models a realistic setting where the feature distributions evolve over time (e.g., due to changes in software frameworks, or malware evolution), while the prediction targets, \ie malware/goodware, remain fixed.
We conduct our DIL experiments on two Android datasets: \elsa, and \tesseract~\cite{pendlebury2019tesseract}.~\footnote{\url{https://ramd-competition.github.io/}} 
Both datasets contain applications collected from the AndroZoo repository~\cite{allix2016androzoo}, labeled as malicious if detected by at least $p$ VirusTotal scanners (\elsa: $p=10$, \tesseract: $p=4$), and as benign if undetected ($p=0$).~\footnote{\label{virustotal} \url{https://www.virustotal.com/gui/home/upload}}
Samples with detections in the $(1, p-1)$ range are discarded.
\elsa includes \nappselsa apps (\ngoodwareelsa goodware, \nmalwareelsa malware), spanning 2017–2019. \tesseract spans 2014–2018 and contains \nappstes apps (\ngoodwaretes goodware, \nmalwaretes malware). 
Following~\cite{pendlebury2019tesseract}, we use a \NMonthsPerSplitelsa-month window per experience, splitting each into 80\% training and 20\% testing, and repeat experiments with 5 random seeds.
\elsa contains \NAppsPerSplitelsa apps per experience (\NAppsTrainPerSplitelsa for training); \tesseract varies in size across experiences, reflecting real-world data imbalance and making it more challenging.
Binary Drebin features~\cite{arp2014drebin} are extracted from the first training split, encoding the presence or absence of specific characteristics from each \texttt{APK}. 
Additionally, for \elsa we extract APIGraph~\cite{zhang2020enhancing} features, a graph-based API representation that extends Drebin by modeling relations among API calls.
For both feature sets, we remove features with variance below $10^{-3}$ as in~\cite{rahman2022limitations}, resulting in \nfeatureselsa and \nfeaturestes Drebin features for \elsa and \tesseract, respectively, and \nfeaturesapig APIGraph features for \elsa.

\myparagraph{CIL Scenario.}
For CIL, we rely on the \azclass dataset~\cite{park2025malcl}, built with the same criteria as \tesseract (\ie APKs collected from AndroZoo, with $p=4$). 
Unlike the DIL setting, the \azclass dataset requires classification among 100 malware families, which is inherently more difficult than the binary case.
The 100 families are split to obtain 10 experiences of 10 classes each. 
For each class, 90\% of the samples are used for training, and the remaining 10\% for testing.
This process is repeated 5 times with randomized class orders.
We extract Drebin features~\cite{arp2014drebin} from the apps and remove those with variance below $10^{-3}$, obtaining \nfeaturesaz features.

\myparagraph{Model Architecture and Training.}
We use a multi-layer perceptron (MLP) as our main classifier, with a hidden layer of size \hiddendim and an output layer of 2 units for DIL and 100 units for CIL.
In the CIL setting, output units corresponding to classes not yet encountered during training are kept inactive throughout the incremental learning process.
We minimize the cross-entropy loss using the SGD optimizer with a learning rate of $10^{-3}$ and momentum of $0.9$.
We also consider a linear support vector machine (SVM) implemented in PyTorch to support incremental training. 
The model minimizes the squared hinge loss using SGD with learning rate $10^{-2}$, momentum $0.9$, and weight decay $10^{-4}$.
We train both models for 30 epochs with a batch size of 32.

\myparagraph{Evaluation Metrics.}
\label{eval-metrics}
For DIL, after training on each experience, we measure the detection performance by considering three metrics: (i) \textit{precision}, (ii) \textit{recall} (\aka detection rate), and (iii) \textit{$F_1$ score}.
To assess security regression, we compute the \textit{NFR} (Eq.~\ref{eq:nfr_cl}), comparing the predictions of the current model and its immediate predecessor, separately for goodware and malware classes.
In the CIL scenario, we measure (i) accuracy, (ii) forgetting, and (iii) NFR.
Following~\cite{chaudhry2018efficient}, after training on the $k$-th experience, we evaluate the metrics on a selected set of test experiences and aggregate the results according to the backward and forward modes, described in \autoref{subsect:sec_reg_cl}.
We obtain a curve for each metric, where each value (one for each test experience) describes the performance after learning the $k$-th experience, reporting its mean and standard deviation across five independent runs.

\myparagraph{CL Methods.}
As our main baselines for comparison, we first consider two strategies: (i) the \naive strategy, \ie finetuning on new experiences without using any knowledge retention mechanism, and (ii) the cumulative strategy, \ie finetuning on both new experiences and the whole data from previous ones, to assess the lower and upper bounds of the performance, respectively.
We then consider five \soa CL methods:
Replay with a buffer size of 200 for DIL experiments, and of 1000 for CIL~\cite{rolnick2019experience}; 
Averaged Gradient Episodic Memory (A-GEM)~\cite{chaudhry2018efficient} with buffer size of 200 for DIL and 1000 for CIL. The number of examples used to compute the reference gradients is set to 32. The number of stored samples per experience is adjusted based on the number of experiences in the dataset.
Elastic Weight Consolidation (EWC)~\cite{kirkpatrick2017overcoming} with $\lambda$=0.001; 
Synaptic Intelligence (SI)~\cite{zenke2017continual} with $\lambda$=0.001 and $\epsilon$=0.1; 
and Learning without Forgetting (LwF)~\cite{li2017learning} with $\alpha$=1 and the temperature=1.
Following~\cite{yan2021positive}, we use PCT with $\alpha=1$, as effective during our grid search, and $\beta$=0.5 showing a reasonable trade-off between NFR reduction and F1-score.
All CL methods are taken from the Avalanche library~\cite{ carta2023avalanche}.

\myparagraph{Evaluating NFRs on Deeper Architectures.}
To complement the experiments on MLP and SVM classifiers, we conduct an additional evaluation in the DIL setting using the CNN-based malware detector proposed in~\cite{mclaughlin2017deep} and implemented in~\cite{liu2026unraveling}.
The detector operates on Android opcode sequences extracted from APK files; we refer the reader to~\cite{mclaughlin2017deep} for further details on the architecture and preprocessing pipeline.
For this experiment, we apply the preprocessing pipeline to the \elsa dataset and truncate opcode sequences to a maximum length of 600,000 elements, padding shorter sequences with zeros.
Models are trained for 10 epochs on each experience.
Due to the computational cost of the experiments, we evaluate only the \naive, cumulative, and PCT strategies.
Results are averaged over three runs with different random seeds.

\subsection{Experimental Results}
\label{subsect:exp_results}
\myparagraph{Measuring Security Regression in CL.}\label{sec:regression_dil}
To evaluate the extent of security regression, we test the considered CL strategies in \textit{backward} mode, \ie measuring performance on past and current data. This evaluation will assess how much of the previously acquired knowledge is retained.
In \autoref{tab:full_backward}, we report the precision, recall, $F_1$, and NFR values for the malware (mw) and goodware (gw) classes under the backward evaluation setting. 
Here, \nfrmalware reflects the rate at which previously detected threats are no longer recognized after an update, while \nfrgoodware indicates the rate at which previously accepted benign samples start being misclassified as malicious.
While all CL strategies maintain high performance on previously seen data, on average, they exhibit high NFR, indicating a considerable amount of security regression. 
For example, on the \elsa dataset, SI and A-GEM reach \nfrmalware of 3.40\% and 3.46\%, exceeding the \naive baseline (3.17\%).  
On \tesseract, NFR values are slightly lower but remain high, with EWC and SI reaching nearly 3\%.  
In both datasets, \nfrgoodware is significantly lower than for malware.
Overall, CL strategies achieve strong precision, recall, and $F_1$ on past and current data, thus mitigating forgetting.  
Yet, security regression remains a consistent issue across methods and datasets.
Additional backward temporal plots are provided in the supplementary material (\autoref{fig:temporal_plots_elsa_backward} and \autoref{fig:temporal_plots_tesseract_backward}).

\myparagraph{Reducing Security Regression in CL.}
\begin{table*}[t] 
\centering 
\rowcolors{2}{gray!20}{white}
\caption{Results on \textbf{\elsa} and \textbf{\tesseract} (\textbf{backward} mode). 
We show two baselines and the CL strategies, alone or combined with PCT, Replay, or both. For each, we measure Precision, Recall, $F_1$, NFR on goodware (gw), and malware (mw).} 
\label{tab:full_backward} 
\setlength{\tabcolsep}{5pt} 
\resizebox{0.99\textwidth}{!}{ 
\input{tables/full_backward}
} 
\end{table*}
We apply the regression-aware PCT penalty to mitigate security regression.
The results in \autoref{tab:full_backward} show that adding PCT to CL strategies reduces regression. 
In particular, SI and A-GEM, whose \nfrmalware values in the \elsa dataset were 3.40\% and 3.46\%, respectively, show a reduction of about 50\% after integrating PCT, reaching 1.19\% and 1.76\%, respectively.
The further addition of a replay buffer further reduces the NFR. 
This is evident in the \tesseract dataset, where SI+Replay+PCT achieves an NFR of 1.43\%, compared to 1.75\% with its exemplar-free version.
This NFR reduction affects the recall but improves precision, resulting in an $F_1$ score that is close to that of CL strategies alone for the \elsa dataset, and only slightly lower for the \tesseract dataset.
In practice, this means the model becomes more consistent in maintaining past correct predictions (especially avoiding new false positives) while slightly reducing detection.
Overall, integrating PCT into CL strategies results in a substantial reduction in NFR, particularly when combined with a replay buffer. 

\myparagraph{Influence of $\alpha$ and $\beta$.}
\begin{figure*}[t]
\centering
\includegraphics[width=0.32\linewidth]{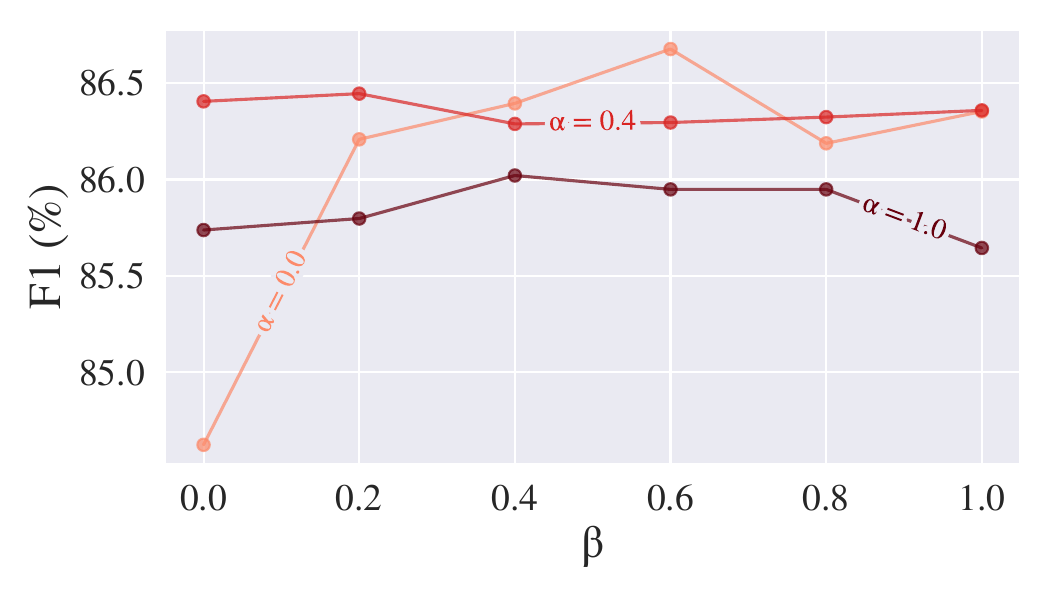}
\includegraphics[width=0.32\linewidth]{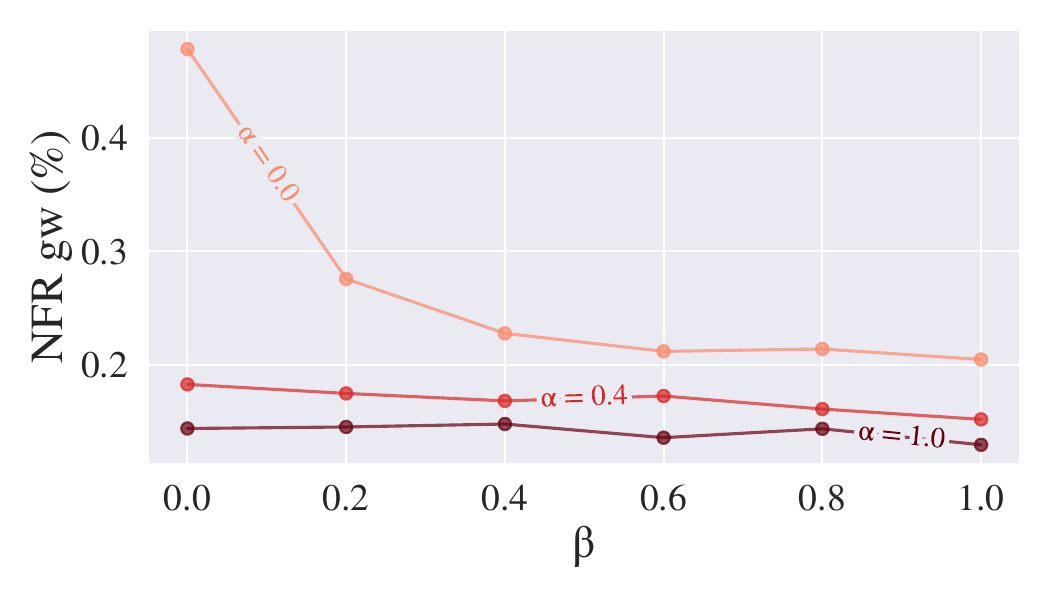}
\includegraphics[width=0.32\linewidth]{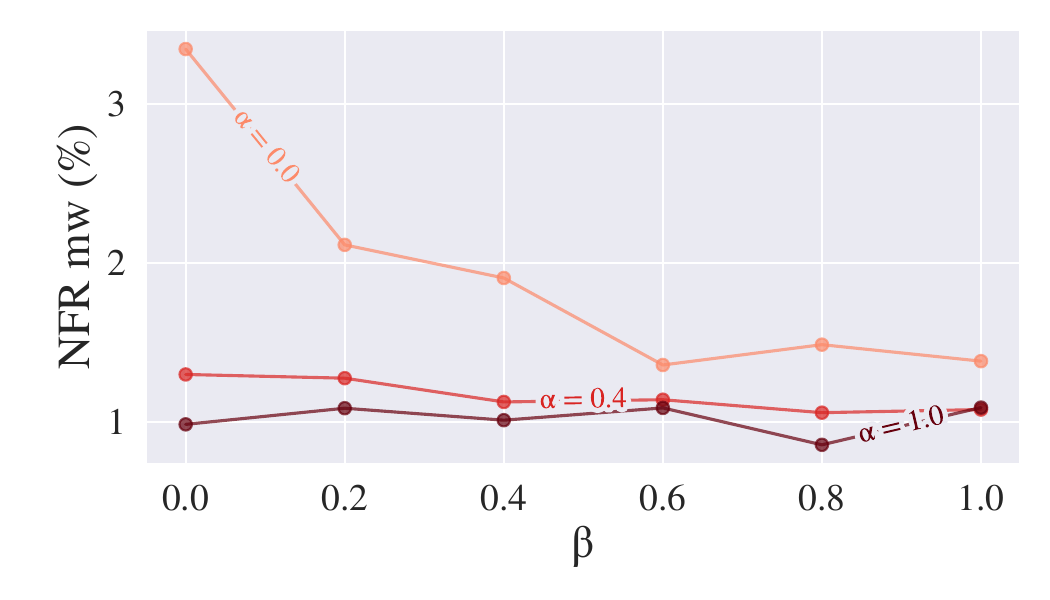}
\caption{Hyperparameter sensitivity of PCT without memory buffer. We show $F_1$ (\textit{left}), NFR for goodware samples (\textit{middle}), and for malware (\textit{right}),  on a grid search over $\alpha$ and $\beta$ in $[0,1]$. Each curve corresponds to a fixed value of $\alpha$, while $\beta$ varies. 
}
\label{fig:PCT_ablation_NB}
\end{figure*}
We analyze the trade-off between classification performance and security regression by investigating the influence of PCT’s $\alpha$ and $\beta$ parameters.
We recall that these are the hyperparameters of PCT, where the scalar $\alpha$ acts as a weighting factor applied to all training samples, ensuring a minimum level of distillation, and the scalar $\beta$ adds extra weight to the distillation loss for samples that were correctly predicted by the old model.
In \autoref{fig:PCT_ablation_NB}, we report the $F_1$ score (left), \nfrgoodware (middle), and \nfrmalware (right) for different configurations of $\alpha$ and $\beta$ in \textit{backward} mode.
For both goodware and malware classes, NFR decreases as $\alpha$ and $\beta$ increase.
The impact of $\beta$ on NFR becomes less significant at higher $\alpha$ compared to lower ones.
Furthermore, the leftmost plot shows that the $F_1$ score also declines as $\alpha$ increases, while the effect of $\beta$ remains minimal, except for small $\alpha$.

\myparagraph{Influence of Buffer Size.}
\begin{figure*}[t]
    \centering
        \includegraphics[width=0.8\linewidth]{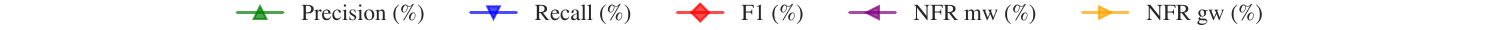}
        \includegraphics[width=0.32\linewidth]{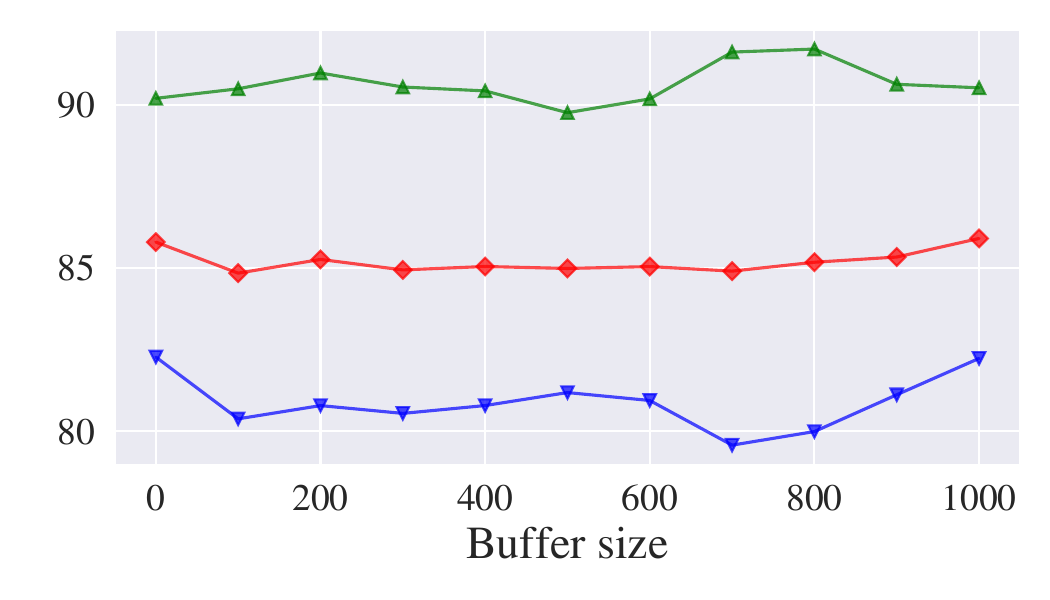}
        \includegraphics[width=0.32\linewidth]{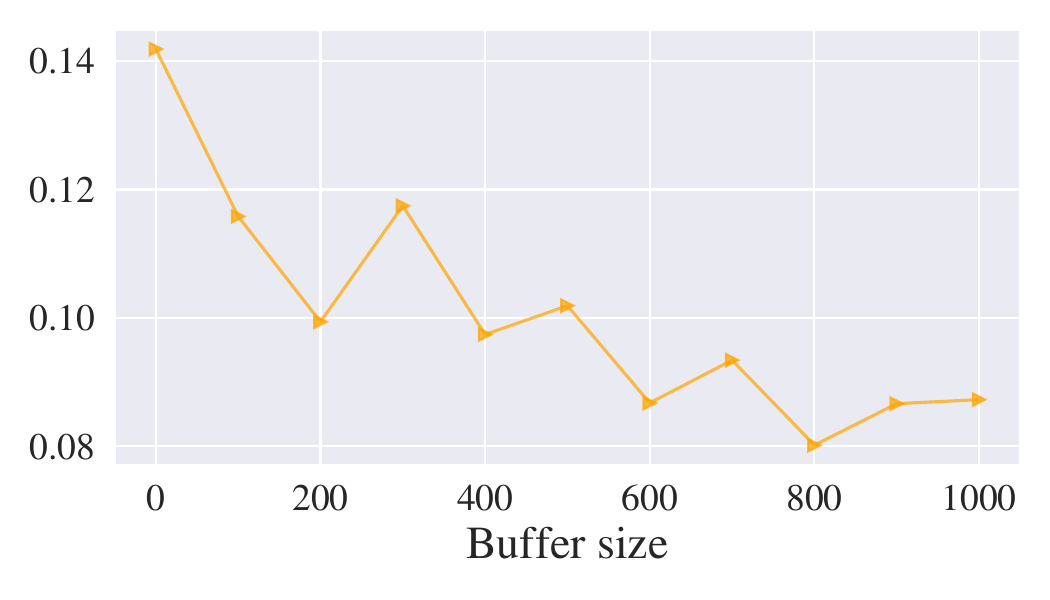}
        \includegraphics[width=0.32\linewidth]{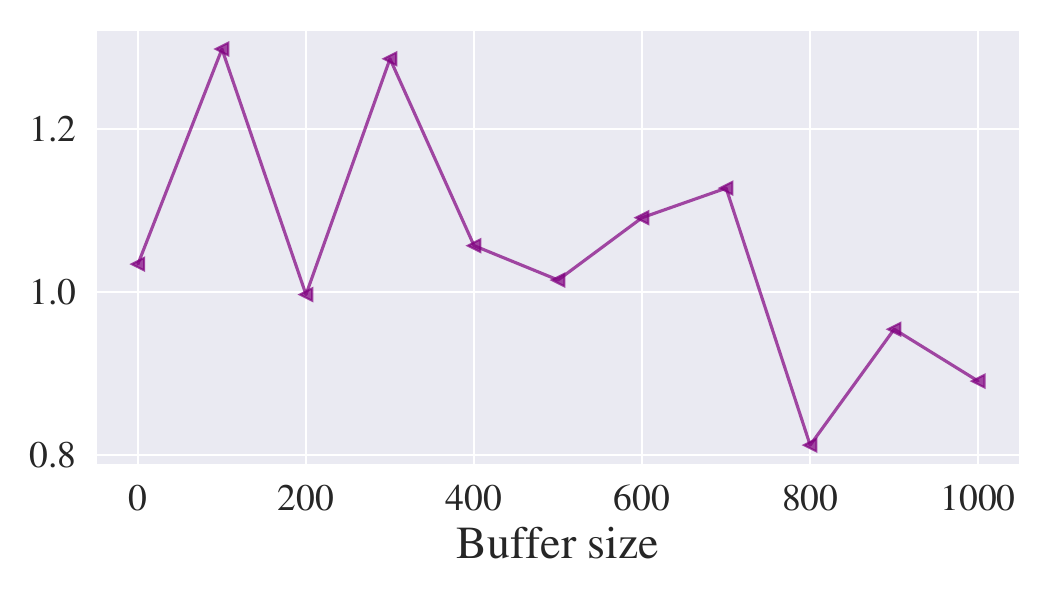}
    \caption{Influence of buffer size, with $\alpha=1$ and $\beta=0.5$. Metrics are shown for different values of buffer size in $[0,1000]$. 
    }
    \label{fig:Buffer_ablation}
\end{figure*}
We now analyze the influence of the buffer size in reducing regression on Replay, keeping $\alpha$ and $\beta$ constant.
We show in \autoref{fig:Buffer_ablation} classification and regression metrics (backward mode) for increasing memory buffer sizes.
Larger buffer sizes significantly reduce regression by nearly 50\% without significantly affecting classification performance.

\myparagraph{Reducing Security Regression in SOTA Detectors.}
\begin{table*}[t] 
\centering 
\rowcolors{2}{gray!20}{white}
\caption{Results on \textbf{ELSA} (\textbf{backward} mode) using \textbf{Linear SVM} and \textbf{MLP} across \textbf{Drebin} and \textbf{APIGraph} feature sets. We show three baselines and the CL strategies, alone or combined with PCT (values in parentheses refer to the +PCT variant, green indicates an improvement, and red a deterioration). For each, we measure $F_1$ and NFR on malware (mw).
}
\label{tab:domain_il_SOTA_backward} 
\setlength{\tabcolsep}{5pt} 
\resizebox{0.99\textwidth}{!}{ 
\input{tables/domain_il_SOTA_backward}
} 
\end{table*}
We measure and mitigate security regression in \soa detectors via additional \elsa experiments using Drebin~\cite{arp2014drebin} and APIGraph~\cite{zhang2020enhancing} feature sets with Linear SVM and MLP models.
In~\autoref{tab:domain_il_SOTA_backward} we report $F_1$ and \nfrmalware  (backward mode).
All strategies across models and feature sets achieve high $F_1$ scores, with the MLP performing slightly better than the Linear SVM. 
However, CL strategies alone exhibit non-negligible \nfrmalware, especially for the Linear SVM, where SI on the APIGraph reaches 6.44\%. 
In comparison, the MLP consistently shows lower regression on both feature sets.
When combining CL strategies with PCT (values in parentheses), predictive performance improves, leading to higher $F_1$ scores across models and feature sets, while the \nfrmalware is substantially reduced. 
For instance, with Linear SVM, \nfrmalware for SI on APIGraph drops from 6.44\% to 1.87\%, highlighting PCT's impact in reducing regression.
Overall, these results show the effectiveness of PCT in improving prediction stability while enhancing predictive performance of \soa detectors.

\myparagraph{Resilience to Concept Drift.}
\begin{table*}[t] 
\centering 
\rowcolors{2}{gray!20}{white}
\caption{
Results on \textbf{\elsa} and \textbf{\tesseract} (\textbf{forward} mode). 
We show two baselines and the CL strategies, alone or combined with PCT, Replay, or both. For each, we measure Precision, Recall, $F_1$, NFR on goodware (gw), and malware (mw).
} 
\label{tab:full_forward} 
\setlength{\tabcolsep}{5pt} 
\resizebox{0.99\textwidth}{!}{ 
\input{tables/full_forward}
} 
\end{table*}
Robustness to concept drift reduces the need for model updates.
We evaluate CL methods in \textit{forward} mode, using future data to measure adaptability to shifting distributions.
In \autoref{tab:full_forward}, we show the precision, recall, $F_1$ score, \nfrmalware, and \nfrgoodware measured in this setting.
While performance drops compared to \textit{backward} mode, CL strategies still adapt to unseen data. 
However, significant regression remains, as seen in \autoref{tab:full_forward}. 
EWC and SI on \elsa reach \nfrmalware of 3.12\% and 3.06\%, respectively, more than double that of the Cumulative baseline (1.46\%).
PCT drastically reduces NFR: EWC+PCT and SI+PCT lower these values to 0.98\% and 0.90\%. 
Adding a replay buffer further improves results, with SI+Replay+PCT achieving 0.57\% malware NFR on \tesseract, down from 0.79\% with SI+PCT.
Even in this case, PCT reduces recall but improves precision, especially in \tesseract, where $F_1$ drops more noticeably.
Detailed forward temporal evolution plots are reported in the supplementary material (\autoref{fig:temporal_plots_elsa_forward} and \autoref{fig:temporal_plots_tesseract_forward}).

\myparagraph{Reducing Regression when Learning Malware Families Incrementally (CIL Scenario).}
\begin{table}[t]
\centering
\rowcolors{5}{white}{gray!20}
\caption{
Results on the \textbf{\azclass} dataset in the \textbf{CIL setting}.
For each strategy, we compute accuracy, forgetting, and NFR, reporting the mean, standard deviation, and worst value.
}
\label{tab:class_il}
\setlength{\tabcolsep}{3pt}
\resizebox{0.99\columnwidth}{!}{
\input{tables/cil_results}
}
\end{table}
%
Beyond detection, family classification is required to better understand and mitigate threats.
In the CIL scenario, at each experience, 10 new families are introduced as output classes; therefore, only \textit{backward} evaluation is applicable.
In this setting, a replay buffer is essential to maintain stability across experiences.
Experiments show that this scenario also suffers from security regression. 
Again, the addition of PCT mitigates the presence of NFs.
\autoref{tab:class_il} shows the results in terms of average accuracy, forgetting, and NFR, reporting the worst case throughout the incremental learning process. 
All CL strategies perform worse than the cumulative baseline, with A-GEM being the worst (51.43\% accuracy, 62.28\% forgetting, 18.94\% NFR).  
Adding PCT improves all strategies across metrics. 
For instance, the accuracy of SI increases from 72.46\% to 78.12\%, while average forgetting and NFR decrease from 29.99\% to 22.16\% and from 9.76\% to 7.42\%, respectively. 
EWC and LwF show similar trends.  
These results show that integrating PCT into CL strategies consistently mitigates performance and regression.

\myparagraph{Evaluating NFRs on a CNN-based Malware Detector~\cite{mclaughlin2017deep}.}
\autoref{tab:DL_backward} reports the results in terms of precision, recall, $F_1$, \nfrmalware, and \nfrgoodware.
The \naive baseline reaches 7.08\% \nfrmalware, while PCT reduces it to 3.35\%, approaching the cumulative baseline (2.91\%).
This reduction comes at the cost of lower recall and $F_1$ score.
Overall, the results confirm that security regression also affects deeper, non-linear model architectures and that it can be mitigated through PCT.

\begin{table}[t]
\centering
\rowcolors{1}{white}{gray!20}
\caption{
Results on the \textbf{\elsa} dataset for the CNN-based malware detector~\cite{mclaughlin2017deep} in the \textbf{DIL setting}.
We show two baselines and the PCT strategy. 
For each, we measure Precision, Recall, $F_1$, and NFR on goodware (gw) and malware (mw).}
\label{tab:DL_backward}
\setlength{\tabcolsep}{3pt}
\resizebox{0.99\columnwidth}{!}{
\input{tables/domain_il_DL_backward}
}
\end{table}

\section{Related Work}
\label{rel}
\myparagraph{Continual Learning for Malware Detection.}
Using CL in malware detection has been first discussed in~\cite{rahman2022limitations}, which provides an extensive analysis of CL methods for malware classification. 
By considering different CL scenarios (DIL, CIL, and TIL), they find replay-based methods useful for malware detection, while regularization-based methods are shown to be less effective. 
%
%
MADAR proposes a novel CL method that employs diversity-aware replay to mitigate catastrophic forgetting~\cite{rahman2025madarefficientcontinuallearning}. 
SSCL-TransMD trains a transformer-based architecture with a replay memory buffer and uses pseudo-labeling~\cite{kou2023sscl}. 
MalFSCIL~\cite{chai2024malfscil} proposes a few-shot CIL approach using decoupled training and a variational autoencoder.  
Further approaches tackle malware detection as continual semi-supervised one-class learning~\cite{chin2024continual}, use multi-modal features~\cite{sun2025temporal}, or propose a generative replay-based approach to create high-quality synthetic malware samples~\cite{park2025malcl}.
However, these studies mainly focus on catastrophic forgetting and overlook the security regression issue. 

\myparagraph{Reducing Regression.} 
Pioneering work in the image classification domain first addressed regression and proposed Positive-Congruent Training (PCT) as a mitigation strategy~\cite{yan2021positive}. This approach has been extended by distilling an ensemble rather than a single model~\cite{zhao2024elodi}, while another work adapts PCT for adversarial robustness~\cite{angioni2024robustnesscongruent}.
Regression has been empirically observed, without the focus on ML, in real-world antivirus systems~\cite{Botacin2020we}, where previously detected malware may become undetected over time.
%
However, analyzing negative flips or integrating regression-aware objectives into the CL pipeline has not been previously explored. 
%
In contrast, our contribution extends CL for malware detection by formally introducing, quantifying, and mitigating security regressions in model updates.
Moreover, our methodology remains applicable to other regression-aware formulations such as~\cite{zhao2024elodi, angioni2024robustnesscongruent}.

\myparagraph{Concept Drift Mitigation}
Several works in malware detection focus instead on anticipating concept drift to mitigate performance degradation on future data.
%
Pendelbury et al.~\cite{pendlebury2019tesseract} first highlighted the impact of temporal evaluation protocols, showing how concept drift can significantly affect malware detectors over time.
Other approaches rely on drift detection mechanisms to trigger updating procedures only when distributional changes are detected~\cite{barbero2020transcending, Xu2019DroidEvolver, Yang2021CADE}, whereas others aim to make models more robust to drift~\cite{angioni2022itasec, Botacin2025Towards}.
In our work, we focus on preserving past decisions rather than adapting to future data distributions, with the update frequency dictated by the CL setting.
Nevertheless, these directions are complementary: drift detection can be used to determine when updates should occur, while our proposed contribution can mitigate forgetting and regression.

\section{Conclusions and Future Work}\label{concl}
In this work, we analyze the phenomenon of security regression in CL applied to malware detection, showing that although many CL strategies limit forgetting, they often fail to ensure consistent predictions after model updates. 
We introduce a regression-aware training framework and instantiate it using the PCT regularizer~\cite{yan2021positive}, demonstrating that it consistently reduces regression with minimal impact on performance across different evaluation modes and scenarios.
We acknowledge that this study assumes clean labels and access to prior model outputs, conditions not always met in real-world deployments. 
Moreover, the \azclass dataset used for multi-class experiments lacks temporal structure, limiting its ability to reflect realistic malware evolution. 
Future work could address these gaps by leveraging active learning to reduce labeling demands or by adopting temporally-aware benchmarks for multi-class settings, better capturing malware dynamics.
We also aim to extend our framework by instantiating it with more recent state-of-the-art regression-aware penalties, such as ELODI~\cite{zhao2024elodi} and RCAT~\cite{angioni2024robustnesscongruent}.
Lastly, our results highlight an important trade-off between preserving past decisions and maximizing average forward performance.
Enforcing regression mitigation is particularly beneficial in security-critical scenarios where reintroducing previously mitigated threats carries a high cost.
Conversely, in highly dynamic environments where malware rapidly becomes obsolete, prioritizing forward performance may be sufficient, making regression mitigation a scenario-dependent design choice.
Despite these limitations, our findings remain broadly applicable and provide a foundation for more robust CL in security-sensitive domains. 
Minimizing regression is key to maintaining trust in continuously updated malware detectors, making them more reliable and scalable in the face of evolving threats.

\section*{Acknowledgments}
This research has been partially supported by the Horizon Europe projects ELSA (GA no. 101070617), Sec4AI4Sec (GA no. 101120393), and CoEvolution (GA no. 101168560); and by SERICS (PE00000014) and FAIR (PE00000013) under the MUR NRRP funded by the EU-NGEU.
This work was carried out while D. Ghiani and S. Gupta were enrolled in the Italian National Doctorate on AI run by the Sapienza Univ. of Rome in collaboration with the Univ. of Cagliari.

\bibliographystyle{ieeetr}
\bibliography{bibliography}

%
%
\begin{IEEEbiography}[{\includegraphics[width=1in,height=1.25in,clip,keepaspectratio]{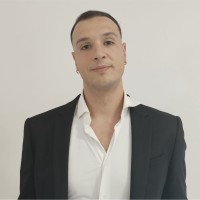}}]{Daniele Ghiani} received his BSc in Computer Engineering (2021) and MSc in Computer Engineering, Cybersecurity, and AI (2024) from the University of Cagliari. He is currently a PhD student in the Italian national PhD programme in AI at Sapienza University, co-located at the University of Cagliari. His research addresses Continual Learning and regression issues in Android malware detection.
\end{IEEEbiography}
\begin{IEEEbiography}[{\includegraphics[width=1in,height=1.25in,clip,keepaspectratio]{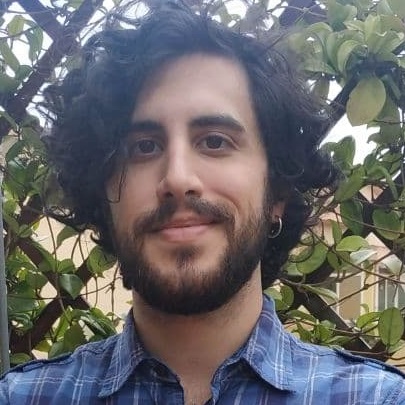}}]{Daniele Angioni} is a Postdoctoral Researcher at the University of Cagliari, Italy. 
He received his PhD in Artificial Intelligence in January 2025 from the Sapienza University of Rome.
His research addresses machine learning security in the real world, applied to both malware and image domains. He serves as a reviewer for top-tier journals, including IEEE TIFS and Pattern Recognition.
\end{IEEEbiography}
\begin{IEEEbiography}[{\includegraphics[width=1in,height=1.25in,clip,keepaspectratio]{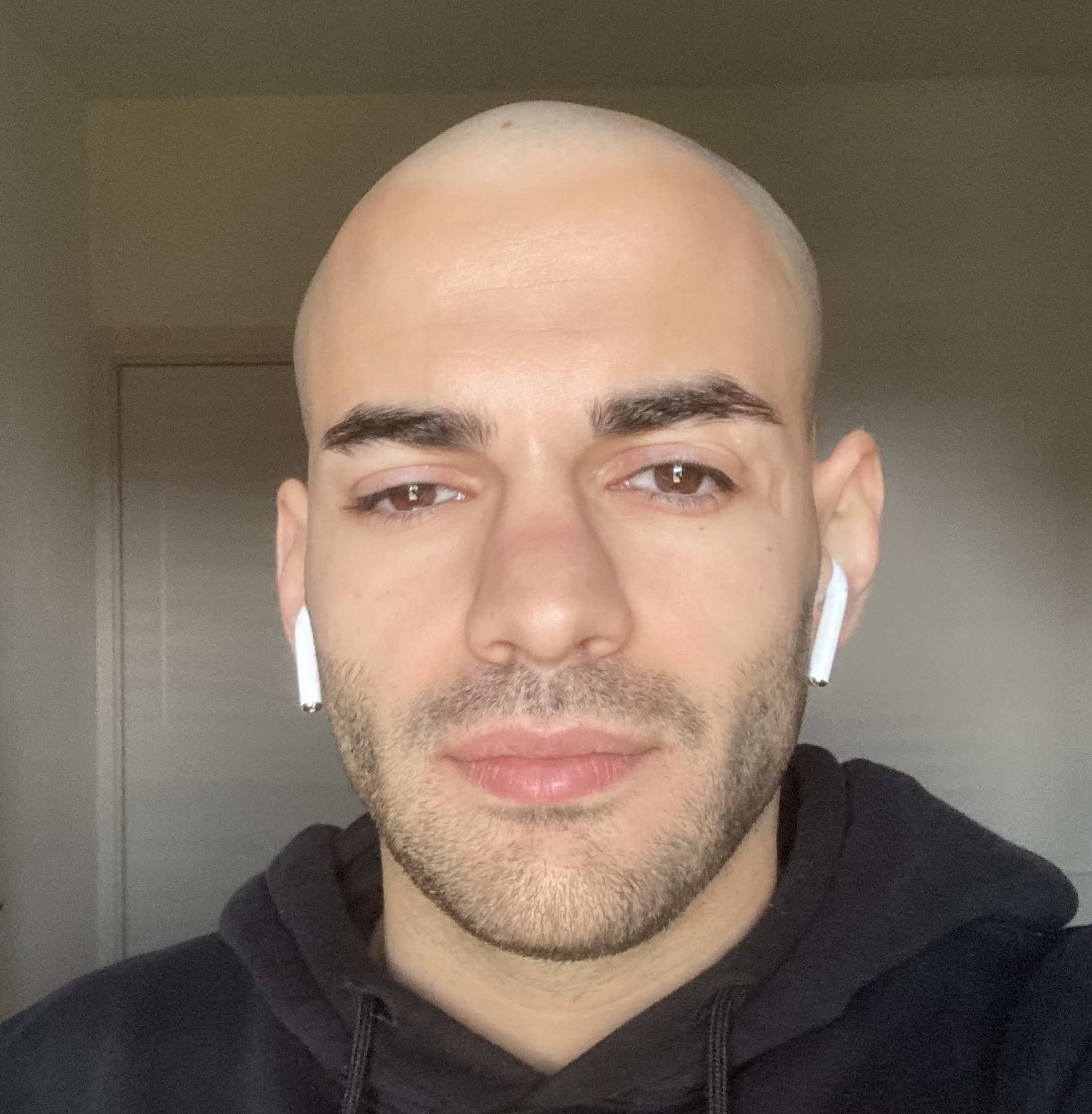}}]{Giorgio Piras} is a Postdoctoral Researcher at the University of Cagliari. He received his PhD in Artificial Intelligence in January 2025 from the Sapienza University of Rome (with honors). His research mainly focuses on adversarial machine learning, with a particular attention to neural network pruning, explainable AI, and LLM security. He serves as a reviewer for journals and conferences, including Pattern Recognition and Neurocomputing journals.
\end{IEEEbiography}
\begin{IEEEbiography}[{\includegraphics[width=1in,height=1.25in,clip,keepaspectratio]{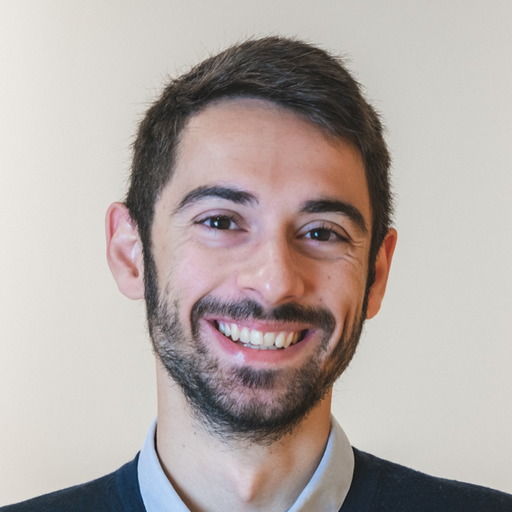}}]{Angelo Sotgiu} is an Assistant Professor at the University of Cagliari. He received from the University of Cagliari (Italy) the PhD in Electronic and Computer Engineering in February 2023. His research mainly focuses on the security of machine learning, also considering practical applications like malware detection. He serves as a reviewer for several journals and conferences.
\end{IEEEbiography}
\begin{IEEEbiography}[{\includegraphics[width=1in,height=1.25in,clip,keepaspectratio]{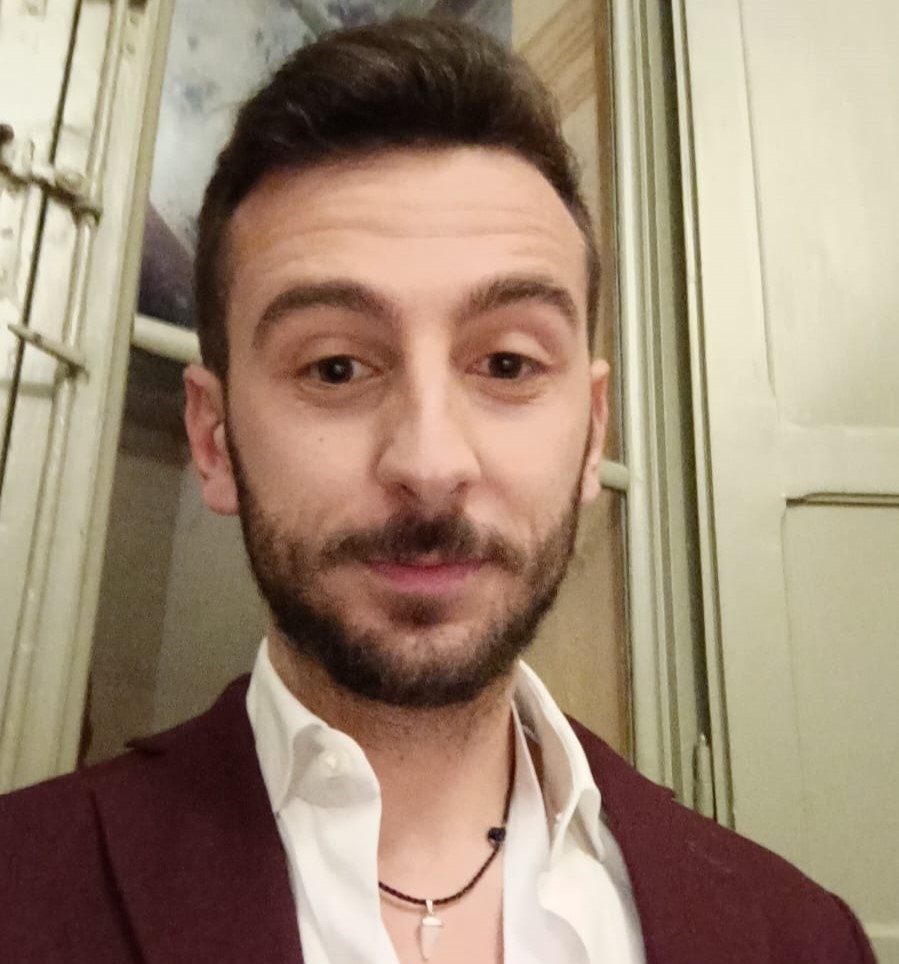}}]{Luca Minnei}is a Ph.D. student in Informatics, Electronics, and Computer Engineering at the University of Cagliari, Italy, he earned a B.Sc. in Computer Science in 2022 and an M.Sc. in Computer Engineering, Cybersecurity, and Artificial Intelligence in 2024, both with honors. His research focuses on malware detection and concept drift in machine learning security.
\end{IEEEbiography}
\begin{IEEEbiography}[{\includegraphics[width=1in,height=1.25in,clip,keepaspectratio]{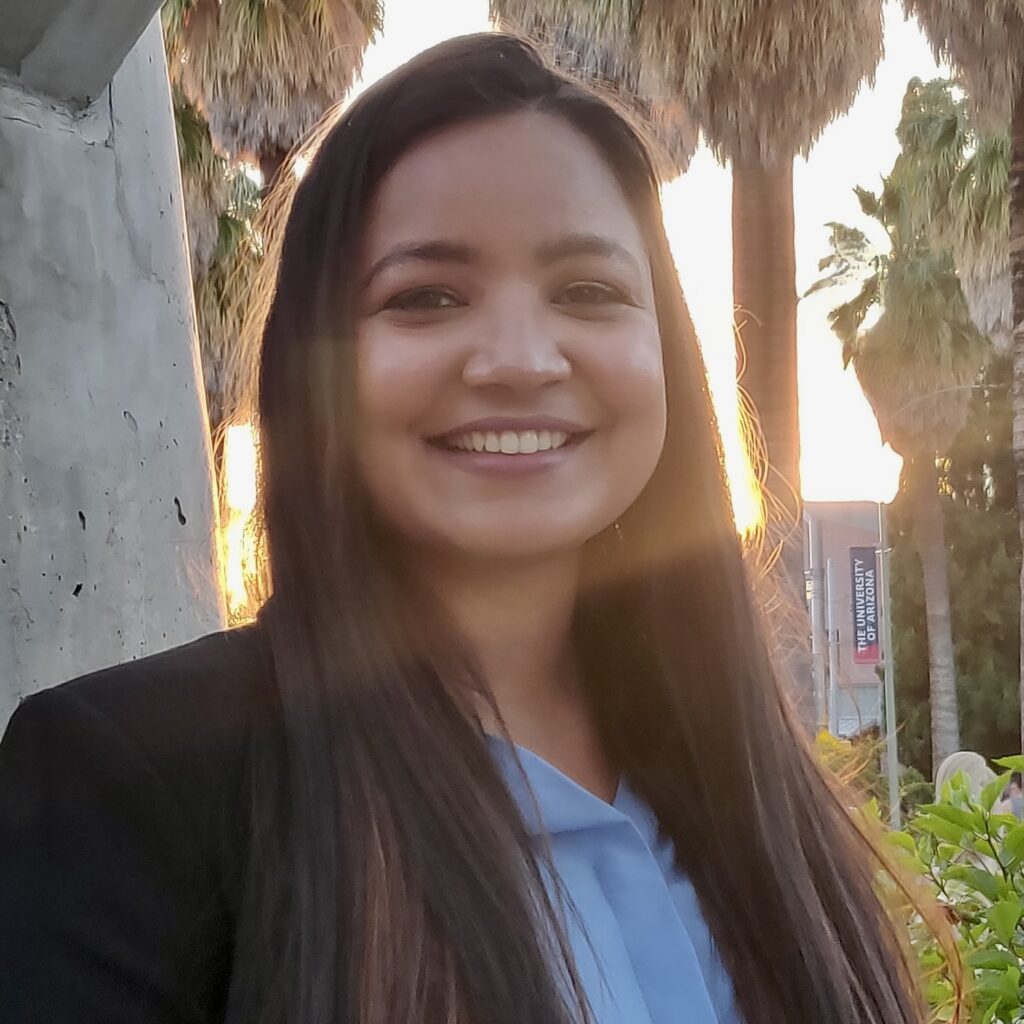}}]{Srishti Gupta} is currently a PhD student in Italian National PhD program in AI at Sapienza University, Rome co-hosted by the University of Cagliari. She received her MS from University of Arizona, US in 2021 and B.Tech from Bharati Vidyapeeth College in Delhi, India in 2017. Her research interests include Continual Learning, Out-of-Distribution Detection and security of LLM models. She serves as a reviewer to several journals and conferences. 
\end{IEEEbiography}
\begin{IEEEbiography}[{\includegraphics[width=1in,height=1.25in,clip,keepaspectratio]{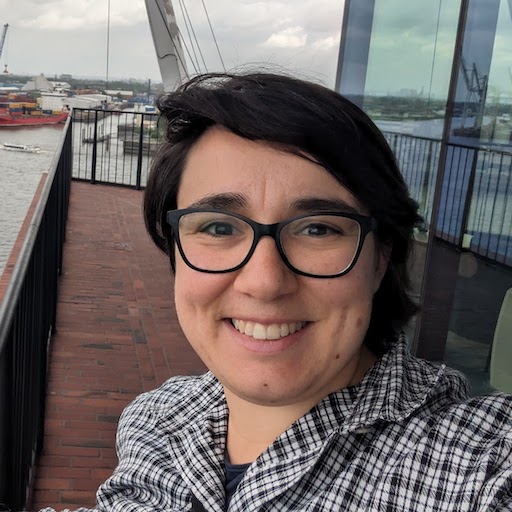}}]{Maura Pintor} is an Assistant Professor at the University of Cagliari, Italy. She received her PhD in Electronic and Computer Engineering (with honors) in 2022 from the University of Cagliari. Her research interests include adversarial machine learning and trustworthy security evaluations of ML models, with applications in cybersecurity. 
She serves as an AC for NeurIPS, and as AE for Pattern Recognition. 
\end{IEEEbiography}
\begin{IEEEbiography}[{\includegraphics[width=1in,height=1.25in,clip,keepaspectratio]{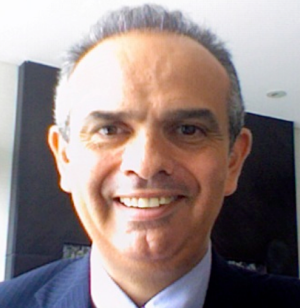}}]{Fabio Roli} Fabio Roli is Full Professor of Computer Engineering at the Universities of Genova and Cagliari, Italy. He is Director of the sAIfer Lab. He is a recipient of the Pierre Devijver Award for his contributions to statistical pattern recognition. He has been appointed Fellow of the IEEE, Fellow of the International Association for Pattern Recognition, Fellow of the Asia-Pacific Artificial Intelligence Association.
\end{IEEEbiography}
\begin{IEEEbiography}[{\includegraphics[width=1in,height=1.25in,clip,keepaspectratio]{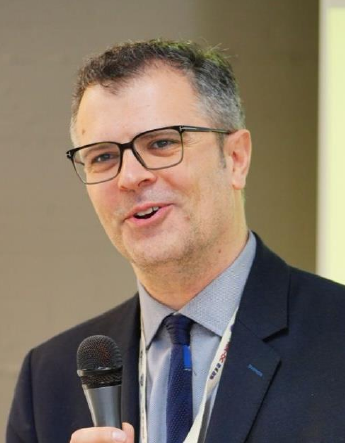}}]{Battista Biggio} is Full Professor of Computer Engineering at the University of Cagliari, Italy. He has provided pioneering contributions to machine learning security. His work on ``Poisoning Attacks against SVMs'' won the 2022 ICML Test of Time Award.  He chaired IAPR TC1 (2016-2020), and served as Associate Editor for IEEE TNNLS and IEEE CIM. He is Associate Editor-in-Chief for Pattern Recognition. He is IEEE Fellow, ACM Senior Member, and Member of IAPR, AAAI, and ELLIS. 
\end{IEEEbiography}







\makeatletter
\def\appendixname{Supplementary Material}
\setcounter{section}{0}
\setcounter{equation}{0}
\setcounter{figure}{0}
\setcounter{table}{0}
\renewcommand{\theequation}{S.\arabic{equation}}
\renewcommand{\thefigure}{S.\arabic{figure}}
\renewcommand{\thetable}{S.\arabic{table}}
\makeatother

\appendix

\renewcommand{\thefigure}{S\arabic{figure}}

\section*{Additional Temporal Plots for Backward Mode Evaluation}
\begin{figure}[H]
    \centering
    \includegraphics[width=0.99\linewidth]{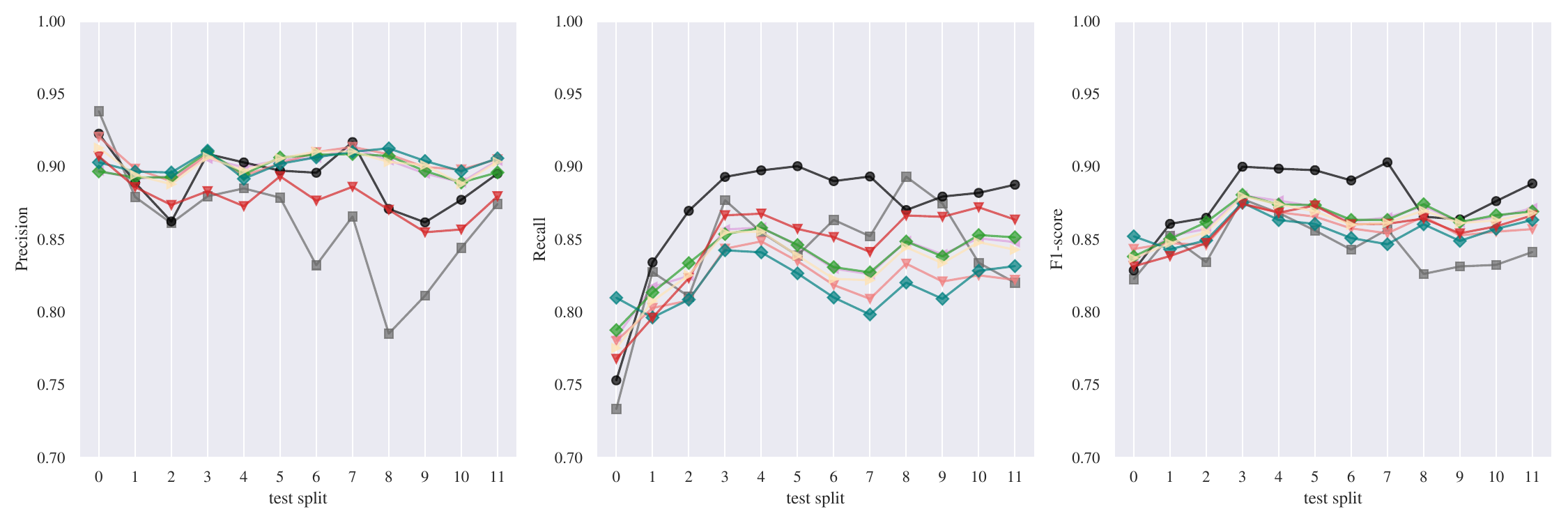}
    \includegraphics[width=0.99\linewidth]{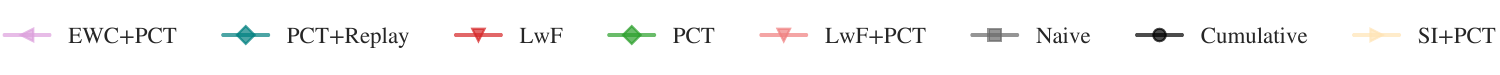}
    \caption{Temporal backward mode performance of different CL methods on the ELSA dataset.}
    \label{fig:temporal_plots_elsa_backward}
\end{figure}

\begin{figure}[H]
    \centering
    \includegraphics[width=0.99\linewidth]{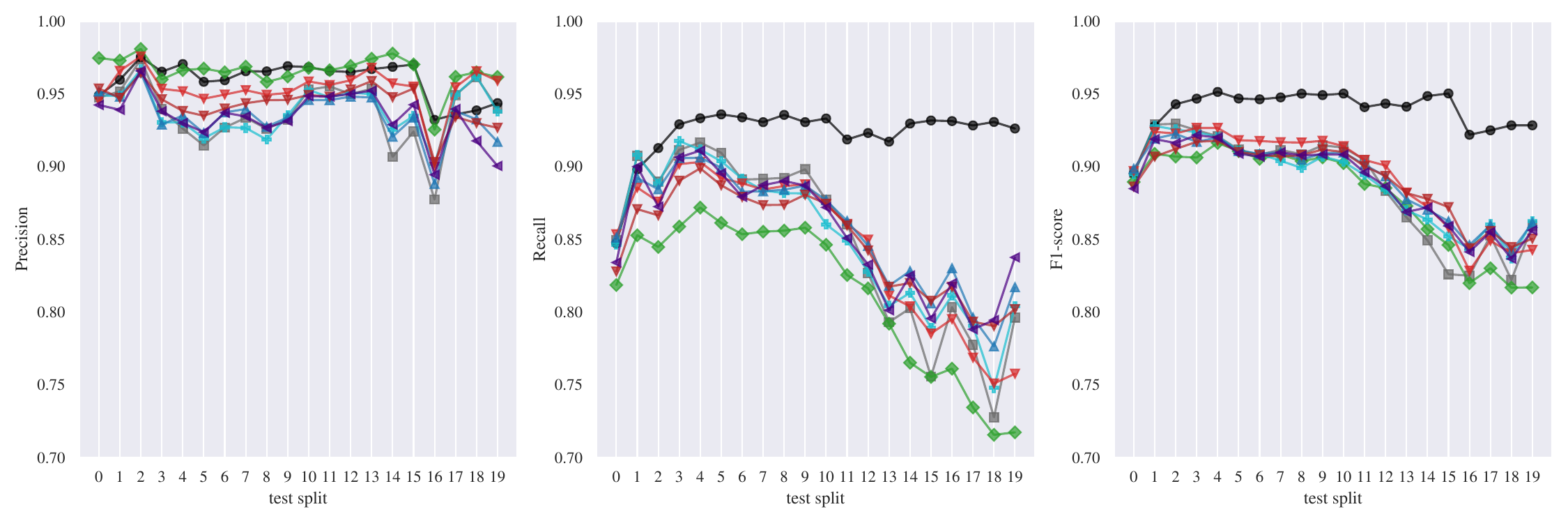}
    \includegraphics[width=0.99\linewidth]{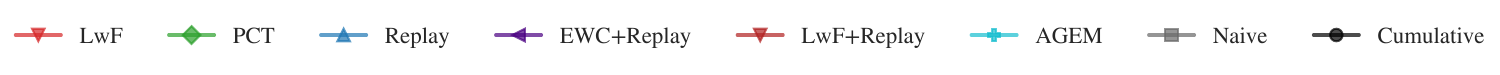}
    \caption{Temporal backward mode performance of different CL methods on the TESSERACT dataset.}
    \label{fig:temporal_plots_tesseract_backward}
\end{figure}

For completeness, we provide temporal backward-mode evolution plots in \autoref{fig:temporal_plots_elsa_backward} and \autoref{fig:temporal_plots_tesseract_backward}, showing, for the ELSA and TESSERACT datasets, the incremental precision, recall, and $F_1$ score of three baselines (cumulative, na\"ive, and PCT) and the five best-performing methods in terms of average $F_1$ score.


\section*{Additional Temporal Plots for Forward Mode Evaluation}
\begin{figure}[H]
    \centering
    \includegraphics[width=0.99\linewidth]{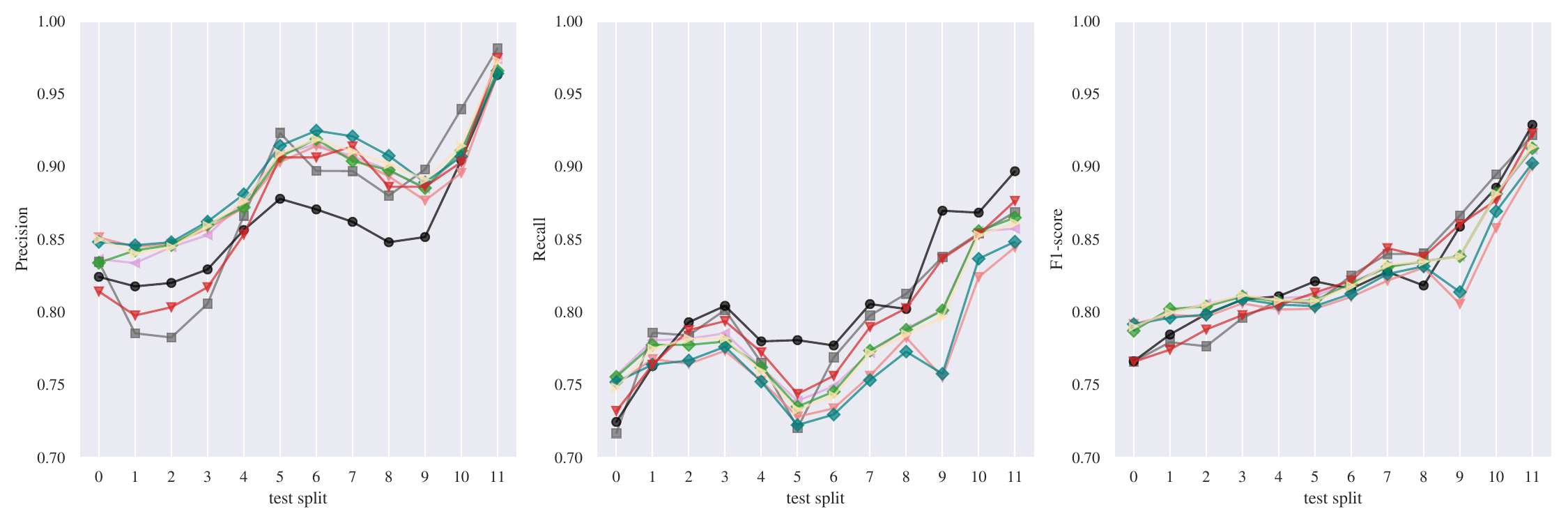}
    \includegraphics[width=0.99\linewidth]{img/prec_rec_f1-clmode-backward-ds-elsa_legend.pdf}
    \caption{Temporal forward mode performance of different CL methods on the ELSA dataset.}
    \label{fig:temporal_plots_elsa_forward}
\end{figure}

\begin{figure}[H]
    \centering
    \includegraphics[width=0.99\linewidth]{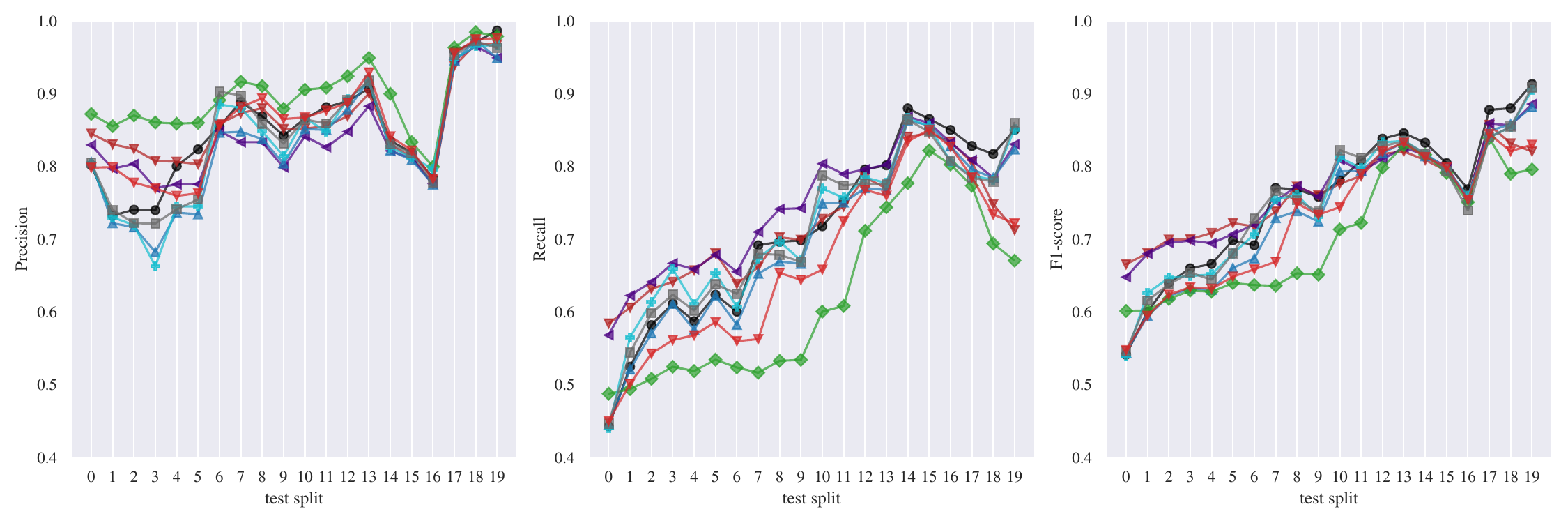}
    \includegraphics[width=0.99\linewidth]{img/prec_rec_f1-clmode-backward-ds-tesseract_legend.pdf}
    \caption{Temporal forward mode performance of different CL methods on the TESSERACT dataset.}
    \label{fig:temporal_plots_tesseract_forward}
\end{figure}

We also provide temporal forward-mode evolution plots in \autoref{fig:temporal_plots_elsa_forward} and \autoref{fig:temporal_plots_tesseract_forward}, showing, for the ELSA and TESSERACT datasets, the incremental precision, recall, and $F_1$ score of three baselines (cumulative, na\"ive, and PCT) and the five best-performing methods in terms of average $F_1$ score.


\end{document}

%% file: commands.tex
\newcommand{\ie}{i.e.,\xspace}
\newcommand{\eg}{e.g.,\xspace}

\newcommand{\aka}{a.k.a.\xspace}

\newcommand{\vct}[1]{\ensuremath{\boldsymbol{#1}}}

\newcommand{\set}[1]{\ensuremath{\mathcal{#1}}}

\newcommand{\myparagraph}[1]{\noindent\textbf{#1}}

\newcommand{\Ind}{\ensuremath{\mathbb{1}}}

\newcommand{\soa}{{state-of-the-art}\xspace}
\newcommand{\naive}{{na\"ive}\xspace}

\newcommand{\elsa}{{ELSA}\xspace}
\newcommand{\tesseract}{{TESSERACT}\xspace}
\newcommand{\azclass}{{AZ-Class}\xspace}

\newcommand{\NMonthsPerSplitelsa}{{3}\xspace}

\newcommand{\nappselsa}{{75,000}\xspace}
\newcommand{\nappstes}{{237,042}\xspace}

\newcommand{\nmalwareelsa}{{7,500}\xspace}
\newcommand{\nmalwaretes}{{26,387}\xspace}
\newcommand{\ngoodwareelsa}{{67,500}\xspace}
\newcommand{\ngoodwaretes}{{232,843}\xspace}
\newcommand{\NAppsPerSplitelsa}{{6,250}\xspace}
\newcommand{\NAppsTrainPerSplitelsa}{{5,000}\xspace}

\newcommand{\nfeatureselsa}{{4,714}\xspace}
\newcommand{\nfeaturestes}{{3,997}\xspace}
\newcommand{\nfeaturesapig}{{4,564}\xspace}
\newcommand{\nfeaturesaz}{{2,439}\xspace}

\newcommand{\hiddendim}{{512}\xspace}

\newcommand{\nfrmalware}{\ensuremath{\text{NFR}_\text{mw}}\xspace}
\newcommand{\nfrgoodware}{\ensuremath{\text{NFR}_\text{gw}}\xspace}

%% file: tables/full_backward.tex
\begin{tabular}{l*{2}{c c c c c}} 
\toprule 
 & \multicolumn{5}{c}{\textbf{ELSA}} & \multicolumn{5}{c}{\textbf{TESSERACT}} \\ 
\cmidrule(lr){2-6} \cmidrule(lr){7-11}
\textbf{Method} & \textbf{Precision (\%)}$\uparrow$ & \textbf{Recall (\%)}$\uparrow$ & \textbf{$F_1$ (\%)}$\uparrow$ & \textbf{\nfrmalware (\%)}$\downarrow$ & \textbf{\nfrgoodware (\%)}$\downarrow$ & \textbf{Precision (\%)}$\uparrow$ & \textbf{Recall (\%)}$\uparrow$ & \textbf{$F_1$ (\%)}$\uparrow$ & \textbf{\nfrmalware (\%)}$\downarrow$ & \textbf{\nfrgoodware (\%)}$\downarrow$ \\ 
\midrule 
Cumulative & $89.17_{\pm 1.93}$ & $87.08_{\pm 3.94}$ & $87.81_{\pm 2.13}$ & $1.16_{\pm 1.00}$ & $0.23_{\pm 0.19}$ & $95.96_{\pm 1.24}$ & $92.28_{\pm 1.95}$ & $93.91_{\pm 1.37}$ & $0.54_{\pm 0.18}$ & $0.07_{\pm 0.05}$\\
Naive & $86.12_{\pm 3.77}$ & $84.00_{\pm 3.99}$ & $84.52_{\pm 1.65}$ & $3.17_{\pm 1.56}$ & $0.48_{\pm 0.36}$ & $93.71_{\pm 2.12}$ & $84.89_{\pm 5.71}$ & $88.78_{\pm 3.59}$ & $2.92_{\pm 2.00}$ & $0.27_{\pm 0.16}$\\
\midrule
Replay & $86.86_{\pm 2.26}$ & $83.19_{\pm 2.93}$ & $84.62_{\pm 1.18}$ & $2.76_{\pm 1.23}$ & $0.36_{\pm 0.18}$ & $93.53_{\pm 1.57}$ & $85.67_{\pm 3.84}$ & $89.22_{\pm 2.56}$ & $2.28_{\pm 1.31}$ & $0.25_{\pm 0.11}$\\
PCT & $89.99_{\pm 0.72}$ & $83.68_{\pm 1.94}$ & $86.48_{\pm 1.10}$ & $1.05_{\pm 0.36}$ & $0.13_{\pm 0.07}$ & $96.58_{\pm 1.09}$ & $81.30_{\pm 5.13}$ & $87.99_{\pm 3.44}$ & $1.75_{\pm 1.22}$ & $0.08_{\pm 0.05}$\\
PCT+Replay & $90.30_{\pm 0.64}$ & $81.86_{\pm 1.49}$ & $85.59_{\pm 0.87}$ & $1.00_{\pm 0.44}$ & $0.10_{\pm 0.05}$ & $96.01_{\pm 1.14}$ & $81.40_{\pm 2.85}$ & $87.89_{\pm 2.20}$ & $1.33_{\pm 0.89}$ & $0.09_{\pm 0.10}$\\
\midrule
LwF & $87.84_{\pm 1.39}$ & $84.49_{\pm 3.15}$ & $85.82_{\pm 1.27}$ & $1.48_{\pm 0.53}$ & $0.23_{\pm 0.11}$ & $95.37_{\pm 1.44}$ & $84.58_{\pm 5.04}$ & $89.38_{\pm 3.23}$ & $1.87_{\pm 1.24}$ & $0.12_{\pm 0.06}$\\
LwF+Replay & $87.35_{\pm 1.48}$ & $83.51_{\pm 2.69}$ & $85.06_{\pm 1.09}$ & $1.29_{\pm 0.62}$ & $0.19_{\pm 0.09}$ & $94.32_{\pm 1.30}$ & $84.87_{\pm 3.47}$ & $89.08_{\pm 2.46}$ & $1.51_{\pm 1.03}$ & $0.15_{\pm 0.12}$\\
LwF+PCT & $90.40_{\pm 0.83}$ & $82.06_{\pm 1.81}$ & $85.73_{\pm 0.92}$ & $1.03_{\pm 0.50}$ & $0.10_{\pm 0.05}$ & $96.67_{\pm 0.96}$ & $79.86_{\pm 4.25}$ & $87.19_{\pm 2.86}$ & $1.55_{\pm 1.06}$ & $0.06_{\pm 0.04}$\\
LwF+Replay+PCT & $89.87_{\pm 0.82}$ & $80.33_{\pm 1.89}$ & $84.50_{\pm 0.98}$ & $1.01_{\pm 0.46}$ & $0.08_{\pm 0.03}$ & $95.74_{\pm 1.29}$ & $80.63_{\pm 2.55}$ & $87.25_{\pm 2.13}$ & $1.09_{\pm 0.67}$ & $0.08_{\pm 0.08}$\\
\midrule
EWC & $86.18_{\pm 3.58}$ & $83.85_{\pm 3.70}$ & $84.49_{\pm 1.59}$ & $3.38_{\pm 1.59}$ & $0.47_{\pm 0.35}$ & $93.80_{\pm 2.08}$ & $84.85_{\pm 5.65}$ & $88.80_{\pm 3.56}$ & $2.93_{\pm 1.99}$ & $0.27_{\pm 0.16}$\\
EWC+Replay & $86.35_{\pm 2.93}$ & $83.64_{\pm 3.65}$ & $84.53_{\pm 1.20}$ & $2.56_{\pm 1.03}$ & $0.39_{\pm 0.22}$ & $93.45_{\pm 1.64}$ & $85.41_{\pm 3.99}$ & $88.92_{\pm 2.72}$ & $2.41_{\pm 1.48}$ & $0.27_{\pm 0.23}$\\
EWC+PCT & $90.03_{\pm 0.68}$ & $83.57_{\pm 2.04}$ & $86.44_{\pm 1.17}$ & $1.21_{\pm 0.46}$ & $0.13_{\pm 0.07}$ & $96.40_{\pm 1.05}$ & $81.50_{\pm 5.23}$ & $88.03_{\pm 3.47}$ & $1.78_{\pm 1.35}$ & $0.08_{\pm 0.05}$\\
EWC+Replay+PCT & $90.00_{\pm 0.73}$ & $81.32_{\pm 1.61}$ & $85.18_{\pm 0.92}$ & $1.11_{\pm 0.42}$ & $0.10_{\pm 0.05}$ & $95.79_{\pm 1.02}$ & $81.38_{\pm 3.08}$ & $87.77_{\pm 2.26}$ & $1.32_{\pm 0.98}$ & $0.07_{\pm 0.05}$\\
\midrule
SI & $85.90_{\pm 3.42}$ & $83.87_{\pm 3.73}$ & $84.38_{\pm 1.66}$ & $3.40_{\pm 1.60}$ & $0.49_{\pm 0.36}$ & $93.80_{\pm 2.06}$ & $85.01_{\pm 5.57}$ & $88.89_{\pm 3.51}$ & $2.92_{\pm 2.03}$ & $0.27_{\pm 0.16}$\\
SI+Replay & $86.52_{\pm 2.16}$ & $82.50_{\pm 3.65}$ & $84.05_{\pm 1.45}$ & $2.82_{\pm 1.47}$ & $0.36_{\pm 0.25}$ & $93.47_{\pm 1.45}$ & $85.08_{\pm 4.24}$ & $88.79_{\pm 2.72}$ & $2.38_{\pm 1.41}$ & $0.26_{\pm 0.14}$\\
SI+PCT & $90.18_{\pm 0.80}$ & $83.09_{\pm 2.18}$ & $86.22_{\pm 1.12}$ & $1.19_{\pm 0.40}$ & $0.13_{\pm 0.06}$ & $96.57_{\pm 1.06}$ & $81.18_{\pm 5.13}$ & $87.91_{\pm 3.42}$ & $1.75_{\pm 1.26}$ & $0.08_{\pm 0.05}$\\
SI+Replay+PCT & $89.95_{\pm 0.83}$ & $81.51_{\pm 2.47}$ & $85.22_{\pm 1.19}$ & $1.16_{\pm 0.59}$ & $0.10_{\pm 0.04}$ & $96.26_{\pm 0.80}$ & $81.31_{\pm 4.02}$ & $87.92_{\pm 2.62}$ & $1.43_{\pm 1.07}$ & $0.05_{\pm 0.02}$\\
\midrule
A-GEM & $86.85_{\pm 2.94}$ & $83.49_{\pm 3.48}$ & $84.73_{\pm 1.08}$ & $3.46_{\pm 1.36}$ & $0.42_{\pm 0.22}$ & $93.82_{\pm 1.57}$ & $85.06_{\pm 4.81}$ & $88.96_{\pm 2.78}$ & $2.83_{\pm 1.49}$ & $0.26_{\pm 0.12}$\\
A-GEM+PCT & $88.55_{\pm 2.15}$ & $81.96_{\pm 1.37}$ & $84.74_{\pm 1.63}$ & $1.76_{\pm 0.94}$ & $0.24_{\pm 0.26}$ & $93.81_{\pm 3.00}$ & $78.04_{\pm 7.20}$ & $84.51_{\pm 6.06}$ & $2.51_{\pm 3.61}$ & $0.18_{\pm 0.15}$\\ 
\bottomrule 
\end{tabular}

%% file: tables/domain_il_SOTA_backward.tex
\begin{tabular}{l  c c c c | c c c c}
\toprule
 & \multicolumn{4}{c|}{\textbf{Linear SVM}} & \multicolumn{4}{c}{\textbf{MLP}} \\
\cmidrule(lr){2-5} \cmidrule(lr){6-9}
 & \multicolumn{2}{c}{\textbf{Drebin}} & \multicolumn{2}{c|}{\textbf{APIGraph}} & \multicolumn{2}{c}{\textbf{Drebin}} & \multicolumn{2}{c}{\textbf{APIGraph}} \\
\cmidrule(lr){2-3} \cmidrule(lr){4-5} \cmidrule(lr){6-7} \cmidrule(lr){8-9}
\textbf{Method} & \textbf{$F_1$ (\%)}$\uparrow$ & \textbf{\nfrmalware (\%)}$\downarrow$ & \textbf{$F_1$ (\%)}$\uparrow$ & \textbf{\nfrmalware (\%)}$\downarrow$ & \textbf{$F_1$ (\%)}$\uparrow$ & \textbf{\nfrmalware (\%)}$\downarrow$ & \textbf{$F_1$ (\%)}$\uparrow$ & \textbf{\nfrmalware (\%)}$\downarrow$ \\
\midrule
Cumulative & 86.48 & 2.38 & 86.45 & 2.06 & 87.81 & 1.16 & 89.48 & 1.11 \\
Naive & 80.73 & 6.13 & 82.20 & 6.36 & 84.52 & 3.17 & 85.13 & 3.95 \\
PCT & 85.63 & 1.71 & 85.07 & 2.00 & 86.48 & 1.05 & 85.67 & 1.59 \\
\midrule
Replay (+PCT) & 82.35 (\textcolor{green!60!black}{85.64}) & 5.24 (\textcolor{green!60!black}{1.28}) & 82.59 (\textcolor{green!60!black}{84.39}) & 5.38 (\textcolor{green!60!black}{1.62}) & 84.62 (\textcolor{green!60!black}{85.59}) & 2.76 (\textcolor{green!60!black}{1.00}) & 84.22 (\textcolor{green!60!black}{84.32}) & 3.38 (\textcolor{green!60!black}{1.22}) \\
LwF (+PCT) & 84.69 (\textcolor{green!60!black}{85.62}) & 2.92 (\textcolor{green!60!black}{1.33}) & 83.83 (\textcolor{green!60!black}{84.75}) & 3.65 (\textcolor{green!60!black}{1.69}) & 85.82 (\textcolor{red}{85.73}) & 1.48 (\textcolor{green!60!black}{1.03}) & 85.48 (\textcolor{red}{85.01}) & 2.23 (\textcolor{green!60!black}{1.29}) \\
EWC (+PCT) & 80.53 (\textcolor{green!60!black}{85.75}) & 6.16 (\textcolor{green!60!black}{1.57}) & 81.98 (\textcolor{green!60!black}{84.82}) & 6.29 (\textcolor{green!60!black}{2.09}) & 84.49 (\textcolor{green!60!black}{86.44}) & 3.38 (\textcolor{green!60!black}{1.21}) & 85.10 (\textcolor{green!60!black}{85.83}) & 3.92 (\textcolor{green!60!black}{1.59}) \\
SI (+PCT) & 80.49 (\textcolor{green!60!black}{85.70}) & 6.10 (\textcolor{green!60!black}{1.74}) & 82.09 (\textcolor{green!60!black}{85.19}) & 6.44 (\textcolor{green!60!black}{1.87}) & 84.38 (\textcolor{green!60!black}{86.22}) & 3.40 (\textcolor{green!60!black}{1.19}) & 85.18 (\textcolor{green!60!black}{85.75}) & 3.86 (\textcolor{green!60!black}{1.54}) \\
AGEM (+PCT) & 81.18 (\textcolor{green!60!black}{84.98}) & 6.02 (\textcolor{green!60!black}{1.60}) & 82.26 (\textcolor{green!60!black}{83.43}) & 6.01 (\textcolor{green!60!black}{2.43}) & 84.73 (\textcolor{green!60!black}{84.74}) & 3.46 (\textcolor{green!60!black}{1.76}) & 84.60 (\textcolor{red}{83.37}) & 4.02 (\textcolor{green!60!black}{2.02}) \\
\bottomrule
\end{tabular}

%% file: tables/full_forward.tex
\begin{tabular}{l*{2}{c c c c c}} 
\toprule 
 & \multicolumn{5}{c}{\textbf{ELSA}} & \multicolumn{5}{c}{\textbf{TESSERACT}} \\ 
\cmidrule(lr){2-6} \cmidrule(lr){7-11}
\textbf{Method} & \textbf{Precision (\%)}$\uparrow$ & \textbf{Recall (\%)}$\uparrow$ & \textbf{$F_1$ (\%)}$\uparrow$ & \textbf{\nfrmalware (\%)}$\downarrow$ & \textbf{\nfrgoodware (\%)}$\downarrow$ & \textbf{Precision (\%)}$\uparrow$ & \textbf{Recall (\%)}$\uparrow$ & \textbf{$F_1$ (\%)}$\uparrow$ & \textbf{\nfrmalware (\%)}$\downarrow$ & \textbf{\nfrgoodware (\%)}$\downarrow$ \\ 
\midrule 
Cumulative & $86.05_{\pm 3.96}$ & $80.54_{\pm 4.75}$ & $82.71_{\pm 4.27}$ & $1.46_{\pm 0.90}$ & $0.23_{\pm 0.17}$ & $84.99_{\pm 7.14}$ & $71.11_{\pm 12.29}$ & $75.75_{\pm 9.76}$ & $1.04_{\pm 0.77}$ & $0.47_{\pm 0.64}$\\
Naive & $87.42_{\pm 5.94}$ & $79.28_{\pm 4.52}$ & $82.66_{\pm 4.63}$ & $3.08_{\pm 1.25}$ & $0.34_{\pm 0.31}$ & $84.14_{\pm 7.85}$ & $70.83_{\pm 11.14}$ & $75.10_{\pm 9.26}$ & $2.09_{\pm 1.60}$ & $0.63_{\pm 0.80}$\\
\midrule
Replay & $86.77_{\pm 6.28}$ & $78.75_{\pm 4.12}$ & $82.06_{\pm 4.56}$ & $2.59_{\pm 1.27}$ & $0.31_{\pm 0.27}$ & $82.55_{\pm 8.00}$ & $69.55_{\pm 11.71}$ & $73.73_{\pm 9.58}$ & $1.82_{\pm 1.54}$ & $0.63_{\pm 0.85}$\\
PCT & $88.70_{\pm 3.65}$ & $78.47_{\pm 3.81}$ & $82.81_{\pm 3.45}$ & $0.84_{\pm 0.52}$ & $0.09_{\pm 0.06}$ & $89.63_{\pm 4.70}$ & $61.91_{\pm 11.45}$ & $70.74_{\pm 8.36}$ & $0.76_{\pm 0.58}$ & $0.30_{\pm 0.54}$\\
PCT+Replay & $89.27_{\pm 3.54}$ & $76.93_{\pm 3.61}$ & $82.17_{\pm 3.15}$ & $0.83_{\pm 0.56}$ & $0.07_{\pm 0.06}$ & $87.66_{\pm 4.67}$ & $64.02_{\pm 9.51}$ & $71.53_{\pm 6.84}$ & $0.65_{\pm 0.42}$ & $0.22_{\pm 0.53}$\\
\midrule
LwF & $87.18_{\pm 5.24}$ & $79.24_{\pm 4.23}$ & $82.57_{\pm 4.36}$ & $1.36_{\pm 0.62}$ & $0.17_{\pm 0.15}$ & $85.42_{\pm 6.80}$ & $66.49_{\pm 11.65}$ & $72.72_{\pm 9.05}$ & $1.01_{\pm 0.88}$ & $0.41_{\pm 0.64}$\\
LwF+Replay & $88.06_{\pm 4.01}$ & $78.82_{\pm 4.29}$ & $82.70_{\pm 3.86}$ & $1.08_{\pm 0.52}$ & $0.15_{\pm 0.12}$ & $85.77_{\pm 5.16}$ & $71.61_{\pm 7.76}$ & $76.20_{\pm 5.42}$ & $1.08_{\pm 1.06}$ & $0.34_{\pm 0.53}$\\
LwF+PCT & $88.54_{\pm 3.32}$ & $76.95_{\pm 3.27}$ & $81.86_{\pm 3.01}$ & $0.79_{\pm 0.49}$ & $0.08_{\pm 0.06}$ & $90.33_{\pm 4.59}$ & $59.58_{\pm 11.05}$ & $69.41_{\pm 8.26}$ & $0.63_{\pm 0.44}$ & $0.24_{\pm 0.70}$\\
LwF+Replay+PCT & $89.00_{\pm 3.41}$ & $75.14_{\pm 2.96}$ & $81.01_{\pm 2.84}$ & $0.76_{\pm 0.37}$ & $0.06_{\pm 0.06}$ & $88.63_{\pm 4.29}$ & $62.59_{\pm 8.67}$ & $70.77_{\pm 6.31}$ & $0.57_{\pm 0.39}$ & $0.18_{\pm 0.33}$\\
\midrule
EWC & $87.50_{\pm 5.79}$ & $79.22_{\pm 4.54}$ & $82.68_{\pm 4.66}$ & $3.12_{\pm 1.38}$ & $0.33_{\pm 0.29}$ & $84.04_{\pm 7.70}$ & $70.67_{\pm 11.33}$ & $74.90_{\pm 9.31}$ & $2.11_{\pm 1.66}$ & $0.63_{\pm 0.73}$\\
EWC+Replay & $87.83_{\pm 4.92}$ & $78.78_{\pm 4.74}$ & $82.55_{\pm 4.31}$ & $2.29_{\pm 1.29}$ & $0.28_{\pm 0.25}$ & $83.70_{\pm 5.68}$ & $74.33_{\pm 8.51}$ & $76.71_{\pm 6.52}$ & $1.82_{\pm 1.21}$ & $0.44_{\pm 0.51}$\\
EWC+PCT & $88.67_{\pm 3.90}$ & $78.58_{\pm 3.59}$ & $82.86_{\pm 3.39}$ & $0.98_{\pm 0.55}$ & $0.10_{\pm 0.07}$ & $88.88_{\pm 4.73}$ & $62.14_{\pm 11.33}$ & $70.75_{\pm 8.24}$ & $0.81_{\pm 0.62}$ & $0.29_{\pm 0.54}$\\
EWC+Replay+PCT & $88.02_{\pm 3.69}$ & $76.82_{\pm 3.36}$ & $81.51_{\pm 3.30}$ & $0.91_{\pm 0.35}$ & $0.09_{\pm 0.07}$ & $88.68_{\pm 4.45}$ & $66.94_{\pm 7.26}$ & $74.24_{\pm 5.10}$ & $0.62_{\pm 0.38}$ & $0.26_{\pm 0.52}$\\
\midrule
SI & $87.54_{\pm 5.79}$ & $79.29_{\pm 4.53}$ & $82.73_{\pm 4.56}$ & $3.06_{\pm 1.36}$ & $0.33_{\pm 0.27}$ & $84.12_{\pm 7.75}$ & $70.64_{\pm 11.40}$ & $74.89_{\pm 9.33}$ & $2.14_{\pm 1.64}$ & $0.63_{\pm 0.81}$\\
SI+Replay & $88.08_{\pm 5.08}$ & $78.10_{\pm 4.71}$ & $82.30_{\pm 4.55}$ & $2.52_{\pm 1.20}$ & $0.27_{\pm 0.22}$ & $84.23_{\pm 5.82}$ & $73.67_{\pm 8.17}$ & $76.77_{\pm 6.33}$ & $1.98_{\pm 1.33}$ & $0.49_{\pm 0.63}$\\
SI+PCT & $89.07_{\pm 3.70}$ & $78.23_{\pm 3.80}$ & $82.84_{\pm 3.45}$ & $0.90_{\pm 0.59}$ & $0.09_{\pm 0.08}$ & $89.41_{\pm 4.75}$ & $61.98_{\pm 11.28}$ & $70.79_{\pm 8.25}$ & $0.79_{\pm 0.55}$ & $0.28_{\pm 0.54}$\\
SI+Replay+PCT & $88.63_{\pm 3.47}$ & $76.56_{\pm 3.27}$ & $81.64_{\pm 3.16}$ & $0.90_{\pm 0.48}$ & $0.09_{\pm 0.06}$ & $89.11_{\pm 4.41}$ & $65.35_{\pm 8.55}$ & $73.57_{\pm 5.99}$ & $0.58_{\pm 0.31}$ & $0.25_{\pm 0.57}$\\
\midrule
A-GEM & $86.57_{\pm 6.20}$ & $79.27_{\pm 4.25}$ & $82.28_{\pm 4.57}$ & $3.03_{\pm 1.37}$ & $0.41_{\pm 0.36}$ & $83.39_{\pm 8.26}$ & $71.16_{\pm 10.82}$ & $75.01_{\pm 9.16}$ & $2.36_{\pm 1.96}$ & $0.69_{\pm 0.83}$\\
A-GEM+PCT & $88.38_{\pm 3.55}$ & $76.31_{\pm 3.36}$ & $81.38_{\pm 3.41}$ & $1.33_{\pm 0.62}$ & $0.12_{\pm 0.08}$ & $87.96_{\pm 4.32}$ & $66.04_{\pm 9.81}$ & $72.72_{\pm 6.96}$ & $1.01_{\pm 0.81}$ & $0.28_{\pm 0.47}$\\ 
\bottomrule 
\end{tabular} 

%% file: tables/cil_results.tex
\begin{tabular}{l*{3}{c c}}
\toprule
 & \multicolumn{2}{c}{\textbf{Accuracy (\%)}$\uparrow$} & \multicolumn{2}{c}{\textbf{Forgetting (\%)}$\downarrow$} & \multicolumn{2}{c}{\textbf{NFR (\%)}$\downarrow$} \\
\cmidrule(lr){2-3} \cmidrule(lr){4-5} \cmidrule(lr){6-7}
\textbf{Method} & Avg. & Worst & Avg. & Worst & Avg. & Worst \\
\midrule
 Cumulative & $90.81_{\pm 3.12}$ & $86.89$ & $2.65_{\pm 0.85}$ & $3.51$ & $1.52_{\pm 0.41}$ & $2.10$ \\
 \midrule
 Replay & $77.70_{\pm 11.19}$ & $62.94$ & $22.71_{\pm 9.52}$ & $35.24$ & $7.74_{\pm 2.15}$ & $10.66$ \\
 PCT+Replay & $77.00_{\pm 8.15}$ & $67.25$ & $22.39_{\pm 5.24}$ & $29.06$ & $7.19_{\pm 1.49}$ & $9.93$ \\
 \midrule
 LwF+Replay & $77.00_{\pm 7.82}$ & $66.65$ & $17.02_{\pm 6.71}$ & $25.86$ & $5.21_{\pm 1.29}$ & $7.32$ \\
 LwF+Replay+PCT & $78.56_{\pm 7.44}$ & $68.66$ & $16.12_{\pm 4.78}$ & $22.60$ & $4.89_{\pm 1.13}$ & $6.55$ \\
 \midrule
 EWC+Replay & $73.09_{\pm 10.07}$ & $60.87$ & $29.38_{\pm 6.25}$ & $37.95$ & $9.68_{\pm 1.73}$ & $12.46$ \\
 EWC+Replay+PCT & $77.98_{\pm 8.66}$ & $66.99$ & $22.30_{\pm 5.57}$ & $29.82$ & $7.51_{\pm 1.64}$ & $9.83$ \\
 \midrule
 SI+Replay & $72.46_{\pm 10.07}$ & $60.45$ & $29.99_{\pm 6.17}$ & $38.31$ & $9.76_{\pm 1.56}$ & $12.24$ \\
 SI+Replay+PCT & $78.12_{\pm 8.59}$ & $67.40$ & $22.16_{\pm 5.51}$ & $29.32$ & $7.42_{\pm 1.67}$ & $9.98$ \\
 \midrule
 A-GEM & $51.43_{\pm 20.61}$ & $30.46$ & $62.28_{\pm 11.24}$ & $74.34$ & $18.94_{\pm 4.19}$ & $25.45$ \\
 A-GEM+PCT & $52.04_{\pm 18.88}$ & $30.44$ & $62.94_{\pm 6.68}$ & $74.00$ & $18.87_{\pm 4.73}$ & $28.33$ \\
\bottomrule
\end{tabular}

%% file: tables/domain_il_DL_backward.tex
\begin{tabular}{l c c c c c}
\toprule
\textbf{Method} & \textbf{Precision (\%)}$\uparrow$ & \textbf{Recall (\%)}$\uparrow$ & \textbf{$F_1$ (\%)}$\uparrow$ & \textbf{\nfrmalware (\%)}$\downarrow$ & \textbf{\nfrgoodware (\%)}$\downarrow$ \\
\midrule
 Cumulative & $87.88_{\pm 5.71}$ & $85.21_{\pm 3.79}$ & $85.43_{\pm 5.09}$ & $2.91_{\pm 1.45}$ & $1.05_{\pm 1.60}$ \\
 Naive & $86.42_{\pm 8.30}$ & $76.33_{\pm 5.38}$ & $79.24_{\pm 4.65}$ & $7.08_{\pm 4.61}$ & $1.84_{\pm 3.37}$ \\
 PCT & $92.45_{\pm 1.74}$ & $63.90_{\pm 4.75}$ & $74.34_{\pm 3.06}$ & $3.35_{\pm 2.55}$ & $0.20_{\pm 0.15}$ \\
\bottomrule
\end{tabular}